\documentclass{article}
\usepackage[nonatbib,preprint]{neurips_2020}
\usepackage[utf8]{inputenc} 
\usepackage[T1]{fontenc}    
\usepackage{hyperref}       
\usepackage{url} 
\usepackage{longtable}
\usepackage{booktabs}       
\usepackage{amsfonts}       
\usepackage{nicefrac}       
\usepackage{microtype}
\usepackage{mathtools,nccmath}
\usepackage{mathtools}
\usepackage{subcaption}

\DeclarePairedDelimiter\floor{\lfloor}{\rfloor}
\usepackage{amssymb}
\usepackage{bm}

\usepackage[
  separate-uncertainty = true,
  multi-part-units = repeat
]{siunitx}
\usepackage{amsmath}
\usepackage{enumerate}
\usepackage{url}            
\usepackage{booktabs}       
\usepackage{amsfonts}       
\usepackage{amsmath}
\usepackage{amssymb}
\usepackage{lipsum}
\usepackage{algorithmicx}
\usepackage{algorithm}
\usepackage{algpseudocode}
\usepackage{hyperref}
\usepackage{bbm}
\hypersetup{
    colorlinks=true,
    linkcolor=blue,
    citecolor=blue,
    filecolor=blue,      
    urlcolor=blue,
}
 \usepackage{array,multirow}
 \usepackage{wrapfig}
 \usepackage{float}
\usepackage[colorinlistoftodos, disable]{todonotes}
\usepackage{wrapfig}
\usepackage[font=small]{caption}
\newcommand{\R}{\ensuremath{\mathbb{R}}}
\newtheorem{theorem}{Theorem}
\usepackage[deletedmarkup=sout]{changes}
\allowdisplaybreaks
\title{Self-Reflective Variational Autoencoder}
\author{%
  Ifigeneia Apostolopoulou, Elan Rosenfeld, Artur Dubrawski 
 \\ Machine Learning Department\\
 Carnegie Mellon University\\
 \texttt{\{iapostol,ekr,awd\}@andrew.cmu.edu}
}

\begin{document}

\maketitle
\newcommand{\attentionpleasecomment}[1]{\textcolor{red}{ #1}}
\newcommand{\ificomment}[1]{\textcolor{black}{ #1}}
\newcommand{\corrections}[1]{\textcolor{black}{ #1}}
\newcommand{\final}[1]{\textcolor{black}{ #1}}
\newcommand{\elancomment}[1]{\textcolor{black}{ #1}}

\begin{abstract}
The Variational Autoencoder (VAE) is a powerful framework for learning probabilistic latent variable generative models. However, typical assumptions on the approximate posterior distribution of the encoder and/or the prior, seriously restrict its capacity for inference and generative modeling. Variational inference based on neural autoregressive models respects the conditional dependencies of the exact posterior, but this flexibility comes at a cost: such models are expensive to train in high-dimensional regimes and can be slow to produce samples. In this work, we introduce an orthogonal solution, which we call \emph{self-reflective inference}. By redesigning the hierarchical structure of existing VAE architectures, self-reflection ensures that the stochastic flow preserves the factorization of the exact posterior, sequentially updating the latent codes in a recurrent manner consistent with the generative model. We empirically demonstrate the clear advantages of  matching the variational posterior to the exact posterior---on binarized MNIST, self-reflective inference achieves state-of-the art performance without resorting to complex, computationally expensive components such as autoregressive layers. Moreover, we design a variational normalizing flow that employs the proposed architecture, yielding predictive benefits compared to its purely generative counterpart. Our  proposed modification is quite general and complements the existing literature; self-reflective inference can naturally leverage advances in distribution estimation and generative modeling  to improve the capacity of each layer in the hierarchy.
\end{abstract}
\section{Introduction}
\elancomment{The advent of deep learning has led to great strides in both supervised and unsupervised learning. One of the most popular recent frameworks for the latter is the Variational Autoencoder (VAE), in which a probabilistic encoder and generator are jointly trained via backpropagation to simultaneously perform sampling and variational inference. Since the introduction of the VAE \cite{kingma2013auto}, or more generally, the development of techniques for low-variance stochastic backpropagation of Deep Latent Gaussian Models (DLGMs) \cite{rezende2014stochastic}, research has rapidly progressed towards improving their generative modeling capacity and/or the quality of their variational approximation. However, as deeper and more complex architectures are introduced, care must be taken to ensure the correctness of various modeling assumptions, whether explicit or implicit. In particular, when working with hierarchical models it is easy to unintentionally introduce mismatches in the generative and inference models, to the detriment of both. In this work, we demonstrate the existence of such a modeling pitfall common to much of the recent literature on DLGMs. We discuss why this problem emerges, and we introduce a simple---yet crucial---modification to existing architectures to address the issue.}

Vanilla VAE architectures make strong assumptions about the posterior distribution---specifically, it is standard to assume that the posterior is approximately factorial (see Figure \ref{fig:iid_vae}). More recent research has investigated the effect of such assumptions which govern the variational posterior \cite{wenzel2020good} or prior \cite{wilson2020bayesian} in the context of uncertainty estimation in Bayesian neural networks; in many scenarios, these restrictions have been found to be problematic. A large body of recent work attempts to improve performance by building a more complex encoder and/or decoder with convolutional layers and more modern architectures (such as ResNets \cite{He_2016_CVPR}) \cite{salimans2015markov,gulrajani2016pixelvae} or by employing more complex posterior distributions constructed with autoregressive layers \cite{kingma2016improved, chen2016variational}. \elancomment{Other work} \cite{tomczak2017vae,klushyn2019learning} focuses on refining the prior distribution of the latent codes.

\elancomment{Taking a different approach, hierarchical VAEs \cite{rezende2014stochastic, gulrajani2016pixelvae, sonderby2016ladder, maaloe2019biva, zhao2017learning} leverage increasingly deep and interdependent layers of latent variables, similar to how subsequent layers in a discriminative network \elancomment{are believed to} learn more and more abstract representations. These architectures exhibit superior generative and reconstructive capabilities since they allow for modeling of much richer latent spaces. While the benefits of incorporating hierarchical latent variables is clear, all existing architectures suffer from a modeling mismatch which results in sub-optimal performance: \emph{the variational posterior does not respect the factorization of the exact \ificomment{posterior distribution of the }generative model.}}

\elancomment{In earlier works on hierarchical VAEs \cite{rezende2014stochastic}, inference proceeds bottom-up, counter to the top-down generative process. To better match the order of dependence of latent variables to that of the generative model, later works \cite{sonderby2016ladder,bachman16} split inference into two stages: first a deterministic bottom-up pass which does necessary precomputation for \ificomment{evidence encoding}, followed by a stochastic top-down pass which incorporates the hierarchical latents to form a closer variational approximation to the \ificomment{exact} posterior. Crucially, while these newer architectures ensure that the order of the latent variables mirrors that of the generative model, the overall variational posterior does not match \ificomment{because of the strong restrictions on the variational distributions of each layer}.}
\paragraph{Contributions.}
In this work, we propose to restructure common hierarchical VAE architectures with a series of bijective layers which enable communication between \ificomment{the inference and generative network}, refining the latent representations. Concretely, our contributions are as follows:
\begin{itemize}
    \item We motivate and introduce a straightforward \emph{rearrangement of the stochastic flow of the model} which addresses the aforementioned modeling mismatch. This modification \ificomment{significantly} compensates for the observed performance gap between  models with only simple layers and \ificomment{those with} complex autoregressive networks \cite{kingma2016improved, chen2016variational}.
    \vspace{-0.05in}
    \item We formally prove that this refinement results in \textbf{a hierarchical VAE whose variational posterior respects the precise factorization of the \ificomment{exact} posterior.} To the best of our knowledge, this is the first such architecture to do so without resorting to computationally expensive autoregressive components or making strong assumptions (e.g. diagonal Gaussian) on the distributions of each layer \cite{sonderby2016ladder}---assumptions that result in degraded performance.
    \vspace{-0.05in}
    \item We experimentally demonstrate the benefits of the improved representation capacity of this model, which stems from the corrected factorial form of the posterior. We achieve state-of-the-art perfomance on MNIST among models without autoregressive layers, and our model performs on par with recent, fully autoregressive models such as \cite{kingma2016improved}. \elancomment{Due to the simplicity of our architecture, we achieve these results for a fraction of the computational cost in both training and inference.}
    \vspace{-0.05in}
    \item \ificomment{We design a \textit{hierarchical variational normalizing flow} that deploys the suggested architecture in order to recursively update the base distribution and the conditional bijective transformations. \elancomment{This architecture} significantly improves upon the predictive performance and data complexity of a Masked Autoregressive Flow (MAF) \cite{papamakarios2017masked} on CIFAR-10.}
\end{itemize}
Finally, it should be noted that our contribution is quite general and can naturally leverage \ificomment{recent advances in \elancomment{variational inference} \ificomment{and deep autoencoders } \cite{chen2016variational,kingma2016improved,tomczak2017vae, burda2015importance, dai2019diagnosing,van2016conditional,rezende2018taming} \ificomment{ as well as } \elancomment{architectural improvements to } \ificomment{density estimation }\cite{gulrajani2016pixelvae,dinh2016density,kingma2018glow,durkan2019neural,oord2016pixel,gregor2015draw}}. We suspect that combining our model with other state-of-the-art methods could further improve the attained performance, which we leave to future work.
\section{Variational Autonencoders}
A Variational Autoencoder (VAE) \cite{kingma2013auto,kingma2019introduction} is a generative model which is capable of generating samples $\pmb{x}\in \R^D$ from a distribution of interest $p(\pmb{x})$ by utilizing latent variables $\pmb{z}$ coming from a prior distribution $p(\pmb{z})$. To perform inference, the marginal likelihood should be computed which involves integrating out the latent variables:
\begin{align}
    p(\pmb{x})&=\int p(\pmb{x},\pmb{z})\,d\pmb{z}.
\end{align}
In general, this integration will be intractable and a lower bound on the marginal likelihood is \ificomment{maximized} instead. This is done by introducing an approximate posterior distribution $q(\pmb{z}\mid\pmb{x})$ and applying Jensen's inequality:
\begin{align}
    & \log p(\pmb{x})= \log\int p(\pmb{x}, \pmb{z})\,d\pmb{z} =  \log\int\frac{q(\pmb{z}\mid\pmb{x})}{q(\pmb{z}\mid\pmb{x})} p(\pmb{x}, \pmb{z})\,d \pmb{z} \ge \int q(\pmb{z}\mid\pmb{x}) \log\left[\frac{p(\pmb{x}\mid\pmb{z})p(\pmb{z})}{q(\pmb{z}\mid\pmb{x})}\right]\,d\pmb{z} \nonumber \\
      &\implies \log p(\pmb{x})\ge \mathop{\mathbb{E}_{q(\pmb{z}\mid\pmb{x})}}[\log  p(\pmb{x}\mid\pmb{z})]- D_{KL}(q(\pmb{z}\mid\pmb{x}) \;\|\; p(\pmb{z})) \triangleq\mathcal{L}(\pmb{x};\pmb{\theta},\pmb{\phi}),
\label{eq:elbo}
\end{align}
where $\pmb{\theta}$, $\pmb{\phi}$ parameterize $p(\pmb{x,z}; {\pmb{\theta}})$ and $q(\pmb{z}\mid\pmb{x};{\pmb{\phi}})$ respectively. \ificomment{For ease of notation, we may omit $\pmb{\theta}, \pmb{\phi}$ in the derivations.}
This objective is called \elancomment{the Evidence Lower BOund (ELBO)} and can be optimized efficiently for continuous $\pmb{z}$ via stochastic gradient descent \cite{kingma2013auto, rezende2014stochastic}.
\section{Self-Reflective Variational Inference} 
\elancomment{We now introduce our main contribution: a \ificomment{hierarchical structure} which ensures that the variational posterior matches the factorization of the \ificomment{exact} \todo{``Exact" is definitely wrong here. You could say ``true" or ``generative"} posterior \corrections{ induced by the generative model}. \ificomment{We refer to this architecture as the Self-Reflective Variational Autoencoder (SeRe-VAE).}} \elancomment{Figure \ref{fig:selfreflective_vae} displays the overall stochastic flow of the joint generative and inference networks}.
\subsection{Generative Model}
\elancomment{Our} generative model consists of a hierarchy of $L$ stochastic layers, as in \cite{rezende2014stochastic}. At each layer, latent variables $\pmb{\epsilon}^l$ are first sampled from a prior distribution $p\elancomment{(\pmb{\epsilon}^l;\pmb{\alpha}^l)}$ \corrections{parameterized by $\pmb{\alpha}^l$} and subsequently transformed to latent variables  $\pmb{z}^l$ by a function $f^l$. 
\elancomment{Formally}, the $l$-th \ificomment{generative }layer can be \final{described by}: $\pmb{z}^l \final{=} f^l(\pmb{\epsilon}^l; \pmb{\beta}^l)$. $f^l$ is a deterministic, bijective function \elancomment{whose parameters \corrections{$\pmb{\beta}^l$} are} a function of \elancomment{the previous layer's latent variables:} $\corrections{\pmb{\beta}^l} \triangleq \pmb{\corrections{\beta}}^l(\pmb{z}^{l-1}\corrections{;\bm{c}^l_\beta})$. The data $\pmb{x} \in \R^D $ is generated at the last layer\elancomment{, drawn from} a suitable distribution \corrections{conditioned} on \corrections{the latent variables $\pmb{z}^{1:L}$}: $\elancomment{\pmb{x} \sim} p(\pmb{x}|\pmb{z}^{1:L};{\pmb{\gamma}})$ with $\pmb{\gamma}\triangleq \pmb{\gamma}(\pmb{z}^{1:L} \corrections{;\bm{c}_\gamma})$. \todo{$\beta$ and $\gamma$ are not parameters, they are functions. Technically we should have here instead the parameters which parameterize the functions. I've added ``with a slight abuse of notation" but we may want to consider just fixing the notation.} The joint distribution of the generative model is \elancomment{thus}:
\begin{align}
    \label{eq:generative_factorization}
    p(\pmb{x},\pmb{z}^1,\pmb{z}^2,\dots,\pmb{z}^L)&= \corrections{p(\pmb{z}^{1})} \times \prod_{\corrections{l=2}}^{L}p(\pmb{z}^l \mid \pmb{z}^{l-1}) \times p(\pmb{x} \mid \pmb{z}^{1:L}),
\end{align}
\corrections{where $\pmb{z}^{1:L}\triangleq\{\pmb{z}^{1}, \pmb{z}^{2},\dots,\pmb{z}^{L}\}$. }We further assume a \textit{correspondence} between the prior and the \ificomment{transformational} layers \elancomment{so that the priors are conditioned on earlier} latent factors; \elancomment{we achieve this by defining the prior layer parameterizations as functions of previous latent codes:} $\corrections{\pmb{\alpha}^l}\triangleq \pmb{\corrections{\alpha}}^l(\pmb{z}^{l-1};\pmb{c}^{l}_\alpha)$.
Intuitively, this choice \elancomment{is justified because} the latent variables $\pmb{\corrections{\epsilon}}^{l}$ of layer $l$, conditioned on $\pmb{z}^{l-1}$, will be able to capture features different than those already captured by $\pmb{z}^{1:{l-1}}$, yielding more meaningful latent representations. In \elancomment{Section \ref{sec:Experiments}, we provide empirical results demonstrating the benefits of this modeling choice.} We denote the parameters of the generative model as $\pmb{\theta}\triangleq\{\pmb{c}^{1:L}_\alpha,\pmb{c}^{1:L}_\beta,\pmb{c}_\gamma\}$. In contrast to \cite{rezende2014stochastic}, in our model the layers of the prior \corrections{are not independent but rather they are} conditioned on the latent representations of the previous layers in the hierarchy.
The proposed model \corrections{also} deviates from other hierarchical architectures \cite{gulrajani2016pixelvae,sonderby2016ladder,maaloe2019biva}; in these models the layers of the prior are conditioned upon the previous prior layers and not upon \ificomment{transformational} layers \ificomment{that are shared between the generative and inference model}\corrections{, as we \elancomment{expound upon} in the following subsection}.
\subsection{Inference Network}
The \ificomment{variational }encoder of the \ificomment{ SeRe-VAE} is defined as follows:
\begin{align}
q(\pmb{\epsilon}^1,\pmb{\epsilon}^2,\dots,\pmb{\epsilon}^L \mid \pmb{\mathcal{D}}) &= q(\pmb{\epsilon}^1 \mid \pmb{\mathcal{D}}) \times \prod_{l=2}^{L}q(\pmb{\epsilon}^{l} \mid \pmb{z}^{l-1},\pmb{\mathcal{D}}). \label{eq:variational_posterior}
\end{align}
\todo{\textbf{TODO: I think the whole first sentence can go. The original DLGM is quite outdated and we don't really ``improve" on it any more than other recent hierarchical architectures, no? Unless I've misunderstood this sentence, in which case it really needs to be clarified.}}
Compared to other hierarchical architectures, \elancomment{(e.g., the Independent VAE, Figure \ref{fig:iid_vae})}, in the proposed model the inference layer\corrections{s are} conditioned on the output of the \corrections{preceding} transformational layer\elancomment{--- these components are} shared between the \corrections{generative} and the inference network (see right half of Figure  \ref{fig:selfreflective_vae}). We now provide formal justification for this choice:
\vspace{-0.05in}
\begin{figure}[!ht]
\centering
\minipage{0.49\textwidth}%
\centering
 \includegraphics[width=0.99\linewidth,height=2.2in]{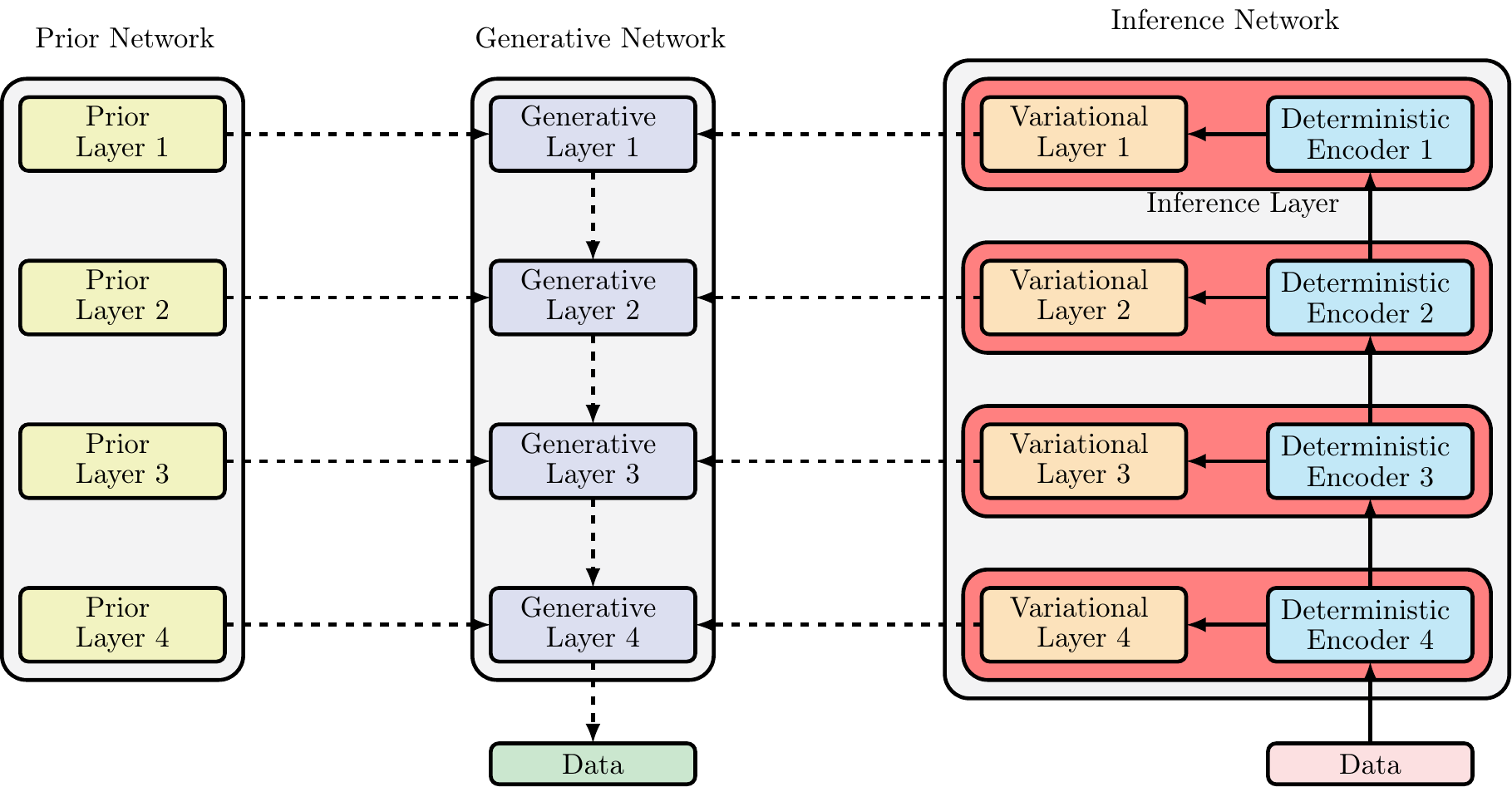}
  \caption{Independent Variational Autoencoder \cite{rezende2014stochastic}}\label{fig:iid_vae}
\endminipage
\minipage{0.49\textwidth}%
\centering
\includegraphics[width=0.99\linewidth,height=2.2in]{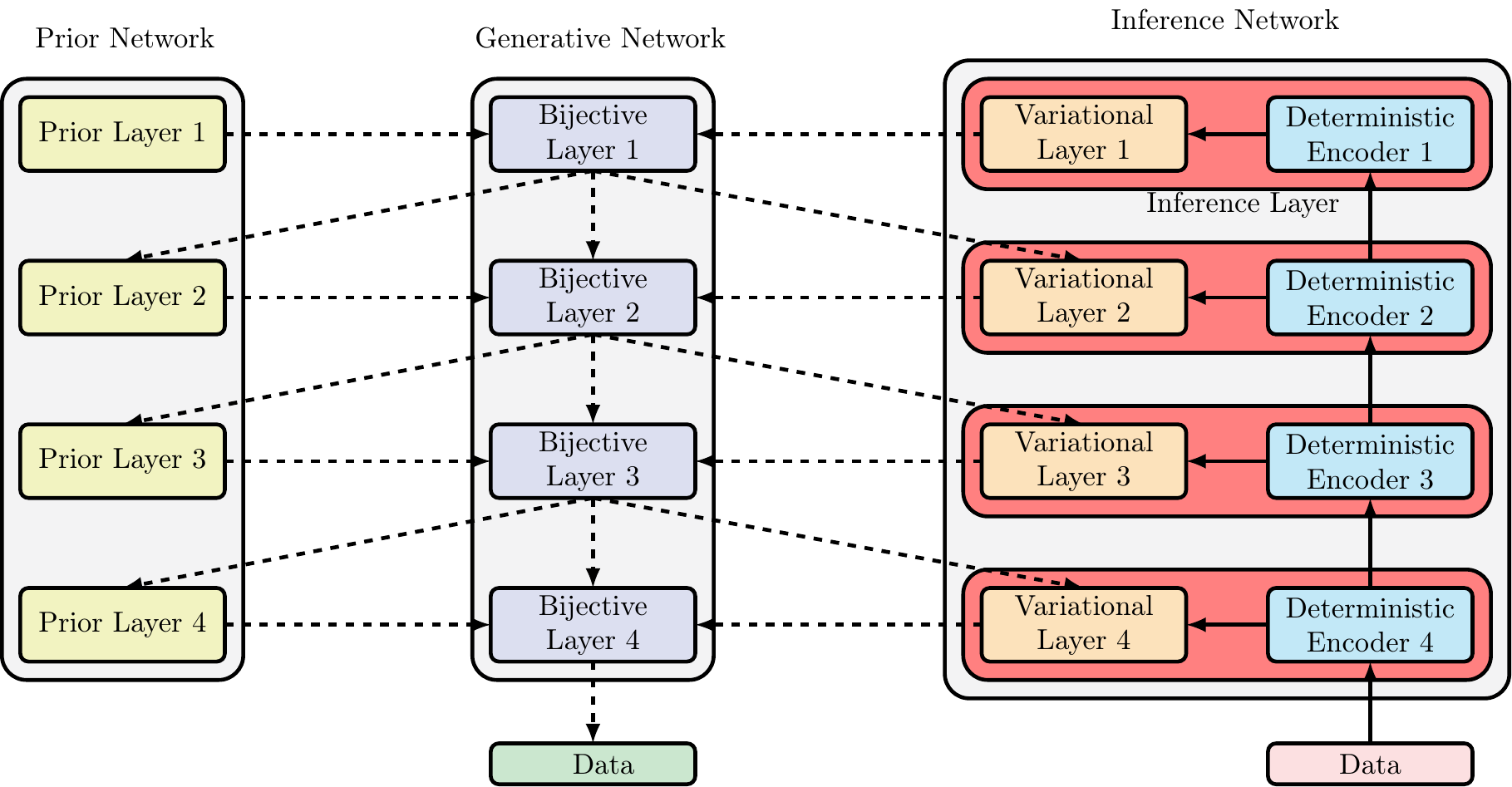}
  \caption{Self-Reflective Variational Autoencoder}\label{fig:selfreflective_vae}
\endminipage
\end{figure}
\vspace{-0.05in}
\begin{figure}[ht]
\centering
\minipage{0.48\textwidth}%
\centering
  \includegraphics[width=0.8\linewidth,height=1.4in]{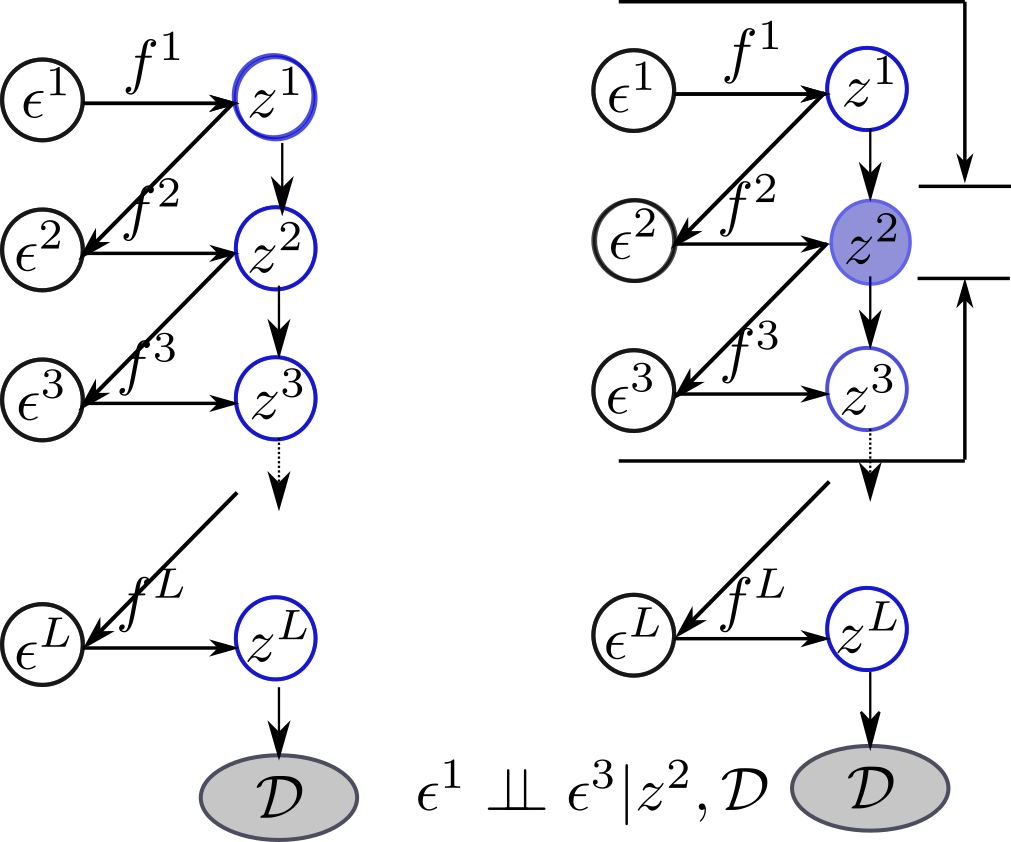}
  \caption{\textit{Generative Model.} The bijectors $f^l$ are conditioned by the latent codes of the previous layer $\pmb{z}^{l-1}: \pmb{z}^l=f^l({\pmb{\epsilon}^l \mid \pmb{z}^{l-1}})$. By the Bayes ball rule, all paths from $\pmb{\epsilon}^3$ to $\pmb{\epsilon}^1$ given $\pmb{z}^2$ and $\mathcal{D}$ are blocked. 
  } \label{fig:self_reflective_gen_model}
\endminipage
\hspace{0.1in}
\minipage{0.43\textwidth}%
\centering
  \includegraphics[width=0.8\linewidth,height=1.4in]{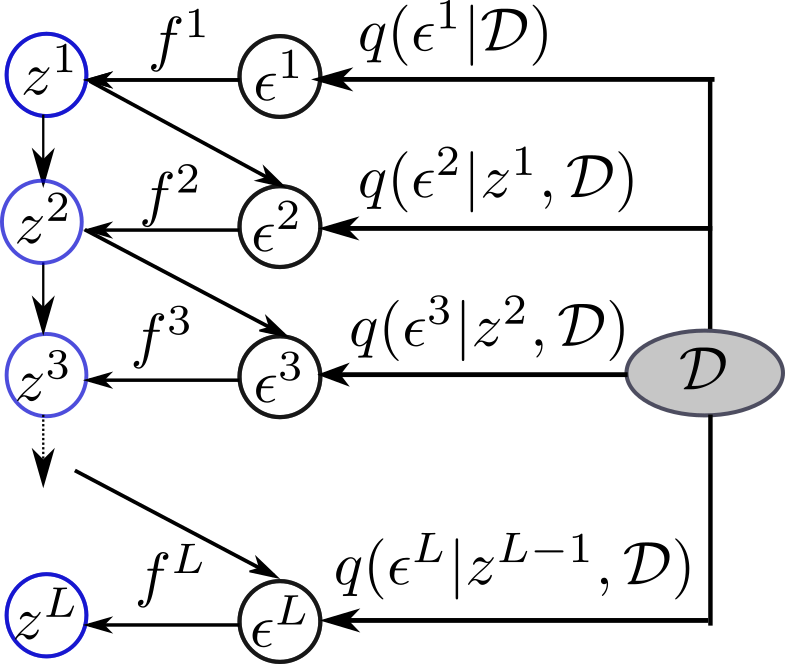}
  \caption{\textit{Variational Inference Model.} The top-down inference enables factorization of the variational factors of the stochastic layers that matches the \ificomment{exact} posterior. 
  }\label{fig:self_reflective_posterior}
\endminipage
\end{figure}
\todo{lettering in Fig 1 and 2 is way too small for being readable}

\begin{theorem} The factorization of the \elancomment{variational posterior defined in \eqref{eq:variational_posterior}} respects the factorization of the \ificomment{exact} posterior distribution \elancomment{induced by \elancomment{the generative model} \eqref{eq:generative_factorization}}.
\end{theorem}
\textbf{Proof:}
Let $p(\pmb{\epsilon}^1,\pmb{\epsilon}^2,\dots, \pmb{\epsilon}^{L} \mid \pmb{\mathcal{D}})$ be the posterior distribution \elancomment{induced by the} generative model \eqref{eq:generative_factorization}, as illustrated in Figure \ref{fig:self_reflective_gen_model}. Then, according to the probability product rule the posterior distribution can be expressed as:
\begin{align}
  p(\pmb{\epsilon}^1,\pmb{\epsilon}^2,\ldots, \pmb{\epsilon}^{L} \mid\pmb{\mathcal{D}}) &= \prod_{l=1}^{L} p(\pmb{\epsilon}^l \mid \pmb{\epsilon}^{<l},\pmb{\mathcal{D}})\corrections{,} \label{eq:posterior_prod}
\end{align}
\corrections{where $\pmb{\epsilon}^{<l}\triangleq\{\pmb{\epsilon}^{1}, \pmb{\epsilon}^{2},\dots,\pmb{\epsilon}^{l-1}\}$. }We will apply the \textit{Bayes ball} rule \corrections{\cite{jordan2003introduction}} to simplify Equation~\eqref{eq:posterior_prod}. \elancomment{Consider an arbitrary layer $l$ of the hierarchy. Because $f^{l-1}$ is a bijector, we have
\begin{align}
    p(\pmb{\epsilon}^l\mid\pmb{\epsilon}^{<l}, \pmb{\mathcal{D}}) &= p(\pmb{\epsilon}^l\mid\pmb{z}^{l-1}, \pmb{\epsilon}^{<l-1},\pmb{\mathcal{D}}).
\end{align}
Now, note that $\pmb{\epsilon}^l$ is \textit{D-separated} from $\pmb{\epsilon}^{l-1}, \ldots, \pmb{\epsilon}^{1}$ by $\pmb{z}^{l-1}$ since all paths from $\pmb{\epsilon}^l$ to $\pmb{\epsilon}^{<l}$  pass through the observed node $\pmb{z}^{l-1}$ (see right half of Figure \ref{fig:self_reflective_gen_model} for an example).
Therefore, we have
\begin{align}
    p(\pmb{\epsilon}^l\mid\pmb{z}^{l-1}, \pmb{\epsilon}^{<l-1},\pmb{\mathcal{D}}) &= p(\pmb{\epsilon}^l\mid\pmb{z}^{l-1},\pmb{\mathcal{D}}).
\end{align}
Since this applies to every layer, it follows that the \ificomment{exact} posterior \eqref{eq:posterior_prod} can also be expressed as
\begin{align}
    p(\pmb{\epsilon}^1,\pmb{\epsilon}^2,\ldots, \pmb{\epsilon}^{L} \mid\pmb{\mathcal{D}}) &= p(\pmb{\epsilon}^1 \mid \pmb{\mathcal{D}}) \times \prod_{l=2}^{L} p(\pmb{\epsilon}^{l} \mid \pmb{z}^{l-1},\pmb{\mathcal{D}}),
\end{align}
exactly matching the factorization \corrections{of the approximate posterior in }\eqref{eq:variational_posterior}.
$\hfill\square$}

From the above analysis, we \elancomment{make some observations about the hierarchy of} \ificomment{ shared transformational} layers in the model:
\begin{itemize}
  \item It allows for complex transformations of the latent variables via the bijective functions $f$.
  \vspace{-0.05in}
  \item It resembles the precision-weighted combination of the generative and inference parts suggested in \cite{sonderby2016ladder}, where in our case the ``averaging" is applied through the shared transformational layers which can \ificomment{express} distributions more complex than diagonal Gaussian.
  \vspace{-0.05in}
  \item By reducing the number of conditioning variables, the \ificomment{ hierarchical transformational layers} offer a convenient way to precisely and efficiently factorize the variational distribution, \ificomment{alleviating the bottleneck present in high-dimensional autoregressive approaches}.
  \item Finally, the use of these layers can be viewed as a hierarchical application of the \textit{reparameterization trick} \cite{kingma2013auto} which is now conducive to a closed-form computation of the KL-divergence since $D_{KL}(q(\pmb{z}\mid\pmb{x}) \;\|\; p(\pmb{z}))= D_{KL}(q(\pmb{\epsilon}\mid\pmb{x}) \;\|\; p(\pmb{\epsilon}))$ (due to bijectivity of $f^l$). The prior and posterior layers can be Gaussian distributions, \elancomment{obviating the need for expensive} \ificomment{Monte Carlo} approximations of the KL divergence \ificomment{ and increasing the stability of the training process}, while the final latent representations of the encoder can be arbitrarily complex.
  \todo{[TODO: SHOULD WE GIVE FORMAL JUSTIFICATION FOR THIS??? KL is invariant of bijective transformations $KL(p||q)==KL(f(p)||f(q)$]}
\end{itemize}


\section{Self-Reflective Normalizing Flows}\label{sec:maf}
\subsection{Definition}
\textit{Normalizing flows} \elancomment{ \cite{tabak2010, tabak2013}, are models} for learning probability distributions based on iterative transformations of samples $\pmb{u}$ drawn from a simple base distribution. Specifically, let $\pmb{x} \in \R^D$ with $\elancomment{\pmb{x} \sim } p(\pmb{x};\pmb{\gamma})$ the distribution of interest. A chain of $T$ invertible transformations $g_{t}$ parameterized by $\pmb{\gamma}_t$ is applied on a sample $\pmb{u}_0 \in \R^D$, drawn from a base distribution $\pi(\pmb{u};{\pmb{\gamma}_0})$ parameterized by $\pmb{\gamma}_0$, such that:

\begin{align}
\pmb{u}_0 \sim \pi(\pmb{u};\pmb{\gamma}_0), \ \pmb{u}_t = g\final{_t}(\pmb{u}_{t-1};\pmb{\gamma}_t) \ \ \forall t=1,\dots,T, \text{and } \pmb{x} \equiv \pmb{u}_T,
\label{eq:norm_flow}
\end{align}

and $\pmb{\gamma}=\{\pmb{\gamma}_{0}, \pmb{\gamma}_{1}, \dots, \pmb{\gamma}_{T}\}$. In the case of invertible and differentiable transformations $g_t$ and differentiable $g^{-1}_t$, the change of variables \elancomment{formula \cite{rudin2006real}} provides a closed-form for $p(\pmb{x})$.
Normalizing flows were popularized for density estimation and variational inference by \cite{dinh2014NICE} and \cite{rezende2015variational}, respectively. An extensive review on normalizing flows is provided in \cite{papamakarios2019normalizing}.
\ificomment{\subsection{Hierarchical Latent Variable Normalizing Flows}\label{sec:sere_maf}
In order to capture high-dimensional dependencies, \elancomment{normalizing flows typically require} a long sequence of transformations $g_t$ and a large hidden dimension, two factors that \elancomment{introduce} scalability issues. Moreover, \elancomment{a generative network} trained jointly with an inference network \cite{kingma2013auto} usually surpasses the predictive capability of purely generative models; see for example Table 2 in \cite{gulrajani2016pixelvae}. These two facts motivate \elancomment{our design of} \textit{ variational normalizing flows.} The latent variables $\pmb{z}$ in this case can be incorporated in the flow in two ways: i) conditioning the base distribution by determining its parameters so that $\pmb{\gamma}_0\triangleq\pmb{\gamma}_0(\pmb{z};\bm{c}^0_\gamma)$, and ii) conditioning the bijective transformations so that $\pmb{u}_t=g_t(\pmb{u}_{t-1}|\pmb{z}\final{;\pmb{\gamma}_t})$.
In the case of a Masked Autoregressive Flow \cite{papamakarios2017masked} or an Inverse Autoregressive Flow \cite{kingma2016improved}, \final{the latter} amounts to designing \textit{conditional} MADE layers \cite{germain2015made} that account for a mask offset so that the additional inputs $\pmb{z}$ are not masked out. We refer to our source code and the supplementary material for the implementation details.\elancomment{ The analogous structure for a hierarchical variational model} is shown in Figure \ref{fig:base_distributions}: each transformation $g_t$ consists of \elancomment{\final{$L$} MADE layers, with each latent code conditioning its respective MADE layer}. For the construction of the base distribution, we are inspired by the residual connections introduced in \cite{He_2016_CVPR}, an adjusted version of which is illustrated in Figure \ref{fig:sere_maf_blocks}. The latent codes $\pmb{z}^l$ combined with a hidden feature map of the previous estimation of the parameters $\pmb{\gamma}_0^{l-1}$ are used to learn a residual mapping $r^l(\pmb{\gamma}_0^{l-1},\pmb{z}^l)$ so that the rectified estimation of the parameters is given by $\pmb{\gamma}_0^{l}=r^l(\pmb{\gamma}_0^{l-1},\pmb{z}^l)+\pmb{\gamma}_0^{l-1}$ and the final estimation by $\pmb{\gamma}_0\equiv\pmb{\gamma}_0^{L}$.}

\begin{figure*}[!htb]
\begin{subfigure}[t]{0.4\textwidth}
\centering
    \includegraphics[width=2in, height=1.9in]{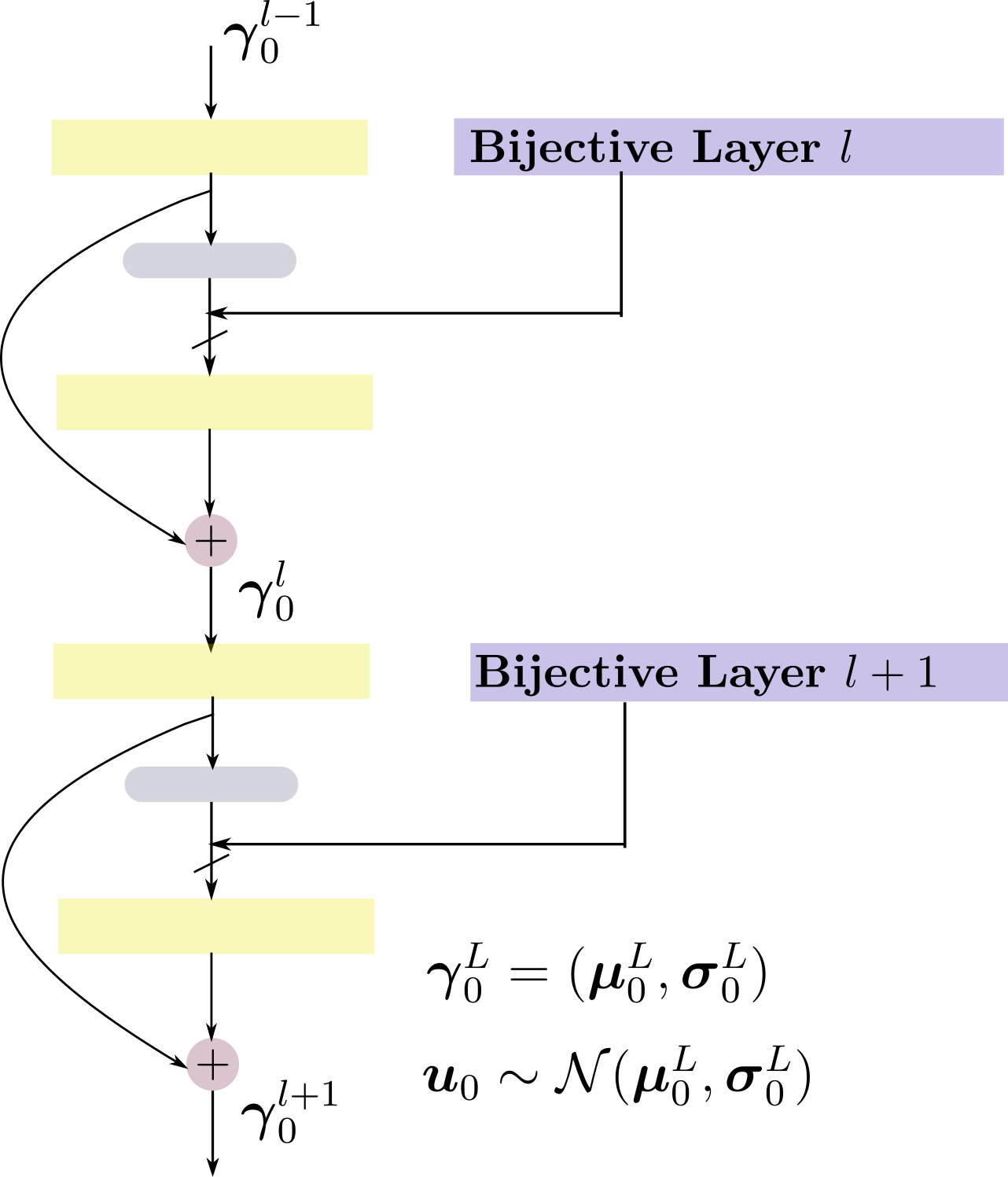}\\
	\caption{\small Building block of \final{a \textit{residual} Gaussian} base distribution}\label{fig:base_distributions}
\end{subfigure}
\hspace{0.5in}
\begin{subfigure}[t]{0.4\textwidth}	
\centering
    \includegraphics[width=2in, height=1.9in]{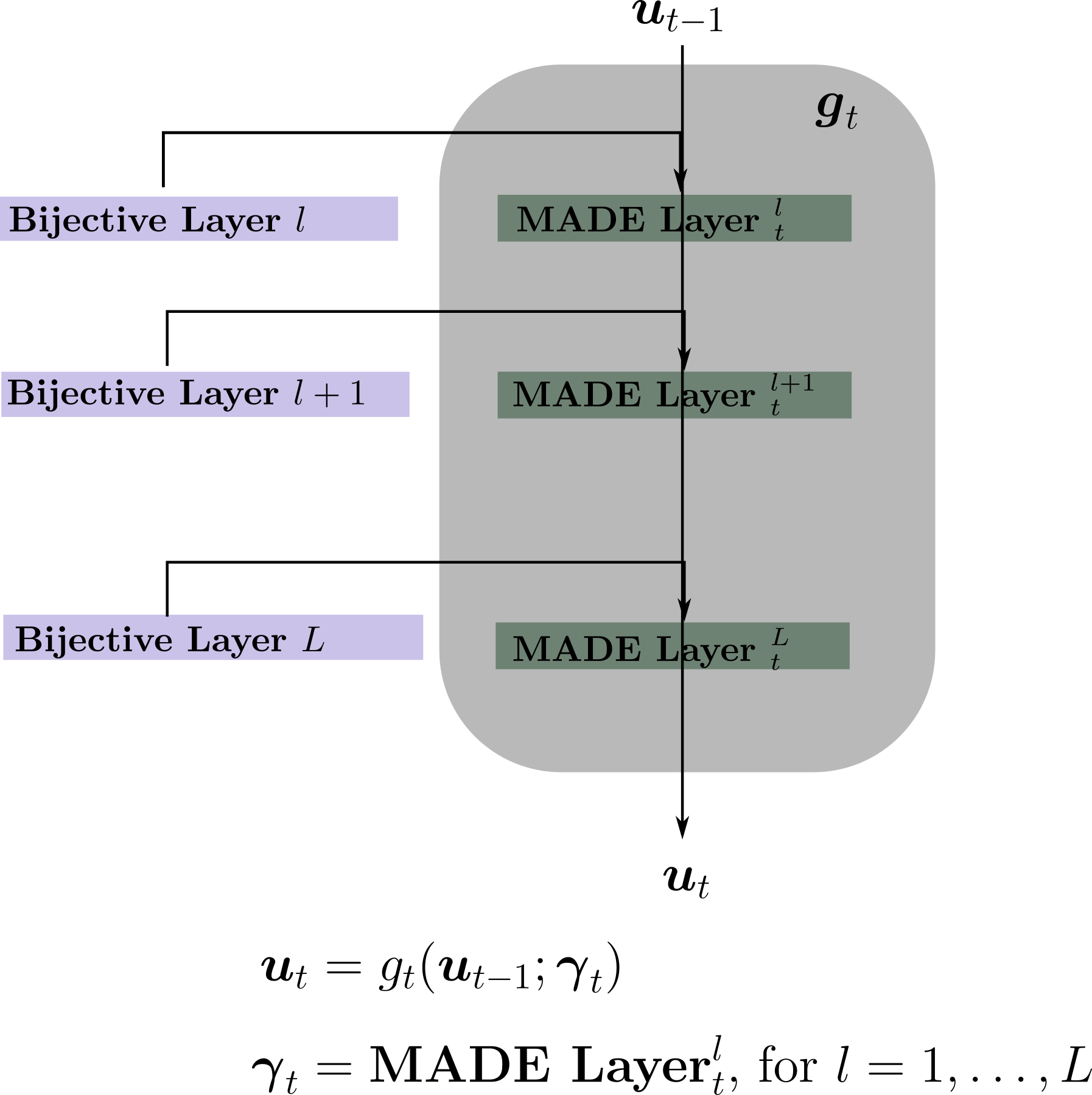} \\
	\caption{\small Building block of the flow $g_t$}\label{fig:bijector}
\end{subfigure}
\caption{Building Blocks of \ificomment{a} Self-Reflective \ificomment{Masked Autoregressive Flow}.}
\label{fig:sere_maf_blocks}
\end{figure*}
\section{Empirical Studies}
\label{sec:Experiments}
\ificomment{Our experiments use}
Tensorflow Probability \cite{tfptensorflow}. All models were optimized using Adam \cite{Kingma2015AdamAM} on a 4xGeForce GTX 2080Ti.
\subsection{Geometric
\todo{[ cite ladder paper and mention that with this scheduling even the IAF model performed better than they are reporting]}
Deterministic Warm-up}\label{sec:warm_up}
\elancomment{For training,} we applied \textit{deterministic warm-up} as suggested in \cite{sonderby2016ladder,rezende2018taming,bowman2015generating,sonderby2016train}, which \elancomment{helps} prevent posterior collapse in deep generative models. This technique introduces a \elancomment{scheduled} regularization coefficient for the \elancomment{KL-divergence term} of the ELBO objective \ificomment{\eqref{eq:elbo}} in order to prevent optimization from being trapped at local optima where the posterior latent representation collapses to the prior distribution. \elancomment{Recall the ELBO objective:}
\begin{equation}
    \mathcal{L}(\pmb{x};\pmb{\theta},\pmb{\phi})= \mathop{\mathbb{E}_q}[\log p(\pmb{x} \mid \pmb{z})]- \beta D_{KL}(q(\pmb{z} \mid \pmb{x}) \;\|\; p(\pmb{z}))  , 0 \le \beta \le 1.
    \label{eq:regularized_elbo}
\end{equation}
Instead of linearly increasing $\beta$, we suggest a schedule that dedicates more optimization steps to larger $\beta$, making the transition to the full ELBO objective smoother. Specifically, \elancomment{for each $n\in\{0,1,\ldots,N\}$, we train for $2^n$ epochs with regularization parameter $\beta_n = 10^{\frac{n}{N}-1}$ (with $\beta_0 = 0$).} \elancomment{We denote with $\mathcal{L}_n$ the objective \eqref{eq:regularized_elbo} evaluated at $\beta=\beta_n$}. \elancomment{Observe that the second \final{phase} of training (\final{after} $2^N$ epochs) is done with the standard VAE objective}. The \elancomment{schedule for} $\beta_n$ for $N=10$ KL levels can be seen in Figure \ref{fig:kl_lambda}.
\subsection{\ificomment{Dynamically binarized MNIST}}
We empirically evaluate the \ificomment{SeRe-VAE} on dynamically binarized MNIST. \ificomment{As in \cite{burda2015importance,sonderby2016ladder,kingma2016improved},} the binary-valued observations are sampled after each epoch with the Bernoulli expectations being set equal to the real\ificomment{, normalized} pixel values in the dataset which prevents overfitting.
\subsubsection{\ificomment{Performance of the MLP SeRe-VAE}}\label{sec: self_refl_mlp_experiments}
\elancomment{To demonstrate that our model's improved performance is due to the restructuring of the stochastic flow and not sophisticated layers, we use simple multilayer-perceptron (MLP) components; we similarly forgo importance weighting \cite{burda2015importance} or any other modeling tricks.} In particular, we use diagonal Gaussian prior and  posterior layers and a \corrections{positive diagonal plus} unit-rank affine transformation for the \elancomment{bijections}. The reader may refer to Equation (19) in \cite{rezende2014stochastic} for the parameterization of this transformation \final{which guarantees positive definiteness and hence invertibility}. \elancomment{We further use} an MLP for the logits of the Bernoulli distribution in the decoder. We adopt a 10-layer architecture, with 10 latent variables \elancomment{per layer, for a} total of 100 latent features being passed to the decoder. Finally, we use \ificomment{independent deterministic encoders }\elancomment{which encourages all latent features to remain active}. \ificomment{The full details of \elancomment{our implementation} are delegated to the supplementary material.} We again emphasize the overall simplicity of our architecture, choosing instead to focus on the benefits of the corrected posterior factorization.
\begin{table}[h]
\centering
\setlength\tabcolsep{0.5pt}
\caption{Dynamically binarized MNIST Performance for VAEs without ResNet layers. 1000 importance samples were used for the estimation of the marginal likelihood. Shown are averages and standard deviations across 5 optimization runs. For the Ladder VAE performance, we refer to Table1 in \cite{sonderby2016ladder}. \ificomment{The models were trained with a single importance sample unless otherwise noted (IW=1).}}
 \resizebox{0.96\textwidth}{!}{
\begin{tabular}[t]{lcc}
\toprule
Model &Details &$log \ p(x) \ge $\\
\midrule
\textbf{Self-Reflective}   &10 layers / 10 variables each, diagonal Gaussian  prior & $\pmb{-81.32 \pm 0.18}$ \\
Importance Weighted Ladder & 5 layers / 128 variables total, $\#$IW samples=10& $-81.74$\\
Ladder & 5 layers / 128 variables total & $-81.84$\\
Self-Reflective IAF  & 10 layers / 10 variables each, Standard Normal Prior & $-81.96 \pm 0.12$\\
 Inverse Autoregressive Flow & 1 layer / 100 variables, Standard Normal Prior & $-83.04 \pm 0.21$ 
 \\
Deep Latent Gaussian Model & 10 layers / 10 variables each, diagonal Gaussian  prior & $-84.53 \pm 0.21$\\
\bottomrule
\label{tab:mnist_marginal}
\end{tabular}}
\end{table}
\begin{figure*}[!h]
\vspace{-0.1in}
\centering
\begin{subfigure}[t]{0.19\textwidth}
\centering
\captionsetup{justification=centering}
    \includegraphics[width=1.5in, height=1.5in]{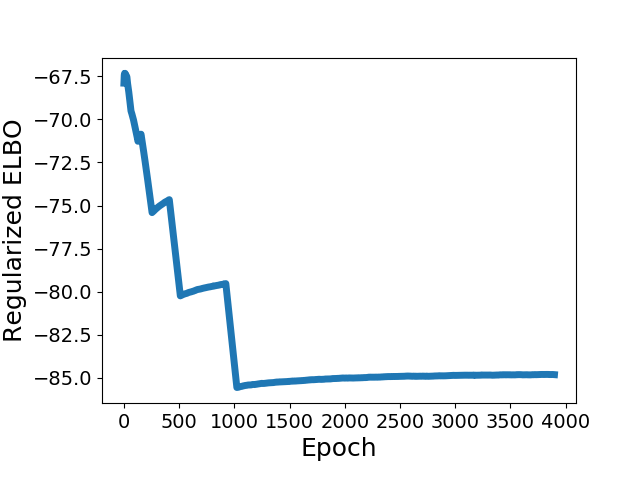}
	\caption{$\mathcal{L}_n $---Eq.~\eqref{eq:regularized_elbo}}
	\label{fig:reg_elbo}
\end{subfigure}
\hspace{0.21in}
\begin{subfigure}[t]{0.2\textwidth}
\captionsetup{justification=centering}
\centering
\includegraphics[width=1.5in, height=1.5in]{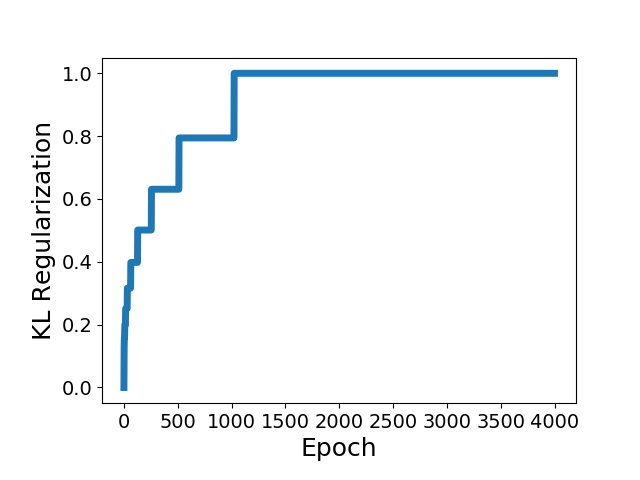}
\caption{$\beta_n$---Eq.~\eqref{eq:regularized_elbo}}\label{fig:kl_lambda}
\end{subfigure}
\hspace{0.21in}
\begin{subfigure}[t]{0.2\textwidth}		
\centering
\captionsetup{justification=centering}
\includegraphics[width=1.5in, height=1.5in]{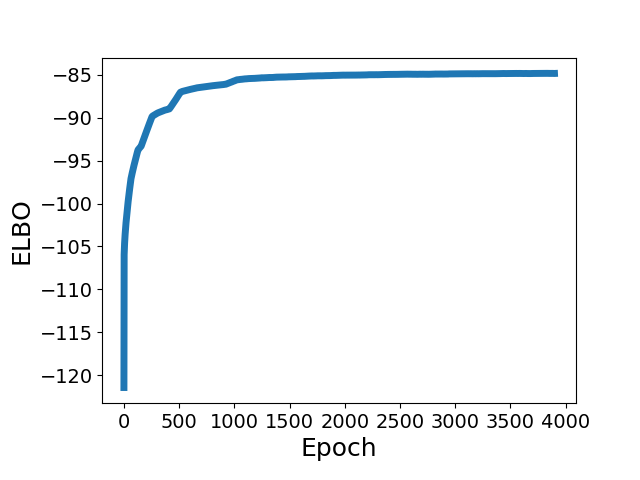}
\caption{ $\mathcal{L}$---Eq.~\eqref{eq:elbo}}\label{fig:elbo}
\end{subfigure}
\hspace{0.21in}
\begin{subfigure}[t]{0.2\textwidth}		
\centering
    \includegraphics[width=1.5in, height=1.5in]{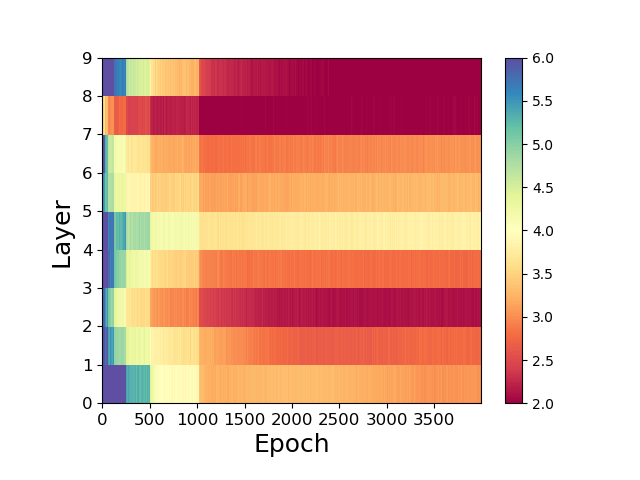}
	\caption{KL per layer}\label{fig:layer_kl}
\end{subfigure}
\caption{Learning curves on validation data for the binarized MNIST dataset.}
\label{fig:reg-overall}
\end{figure*}

\ificomment{As shown in Table \ref{tab:mnist_marginal}, 
our model (SeRe-VAE) outperforms existing models of the same complexity such as the DLGM and Ladder VAE (LVAE), \elancomment{those of} higher complexity such as Inverse Autoregressive Flow (IAF), and models trained with importance weighted samples (IW-LVAE). Note that the architecture of the DLGM is identical to that of SeRe-VAE; to ensure a fair comparison, the DLGM was given larger feature maps in the encoders to compensate for the additional transformational layer inputs in the SeRe-VAE. Therefore, \textit{the performance benefits are \final{solely} attributed to the inclusion of the latent codes in subsequent layers in the hierarchy} (compare Figures \ref{fig:iid_vae} and \ref{fig:selfreflective_vae})}.
\ificomment{Our model outperforms the LVAE models, despite using a smaller latent dimensionality (128 vs. 100) and being trained with \corrections{a single importance sample}.}
\ificomment{Moreover, our model exhibits superior performance compared to the autoregressive IAF; this discrepancy could stem from the 1-layer architecture or the fact that a standard normal prior was used. This result indicates that \final{a prior of equivalent expressive capacity communicating with the generative network could yield additional improvement}. \elancomment{Finally, in our experiments the 10-layer IAF took nearly \emph{twice as long} to train compared to the \ificomment{SeRe-VAE}.} 
} 

\ificomment{\elancomment{Figure \ref{fig:reg-overall} displays learning curves and KL-divergence.} The regularized ELBO in Figure \ref{fig:reg_elbo} is decreasing during the warm-up because the regularization term $\beta_n$ increases during training, see \eqref{eq:regularized_elbo}. Especially in Figure \ref{fig:layer_kl}, we notice that all latent features remain active ($\final{D_{KL}>0}$) and evenly expressive, and no posterior collapse is observed.}
\subsubsection{Performance of the ResNet SeRe-VAE}\label{sec: self_refl_resnet_experiments}
To demonstrate the capacity of our model when combined with complex layers, we replaced the MLPs with ResNets as in \cite{salimans2015markov}. \ificomment{In order to combine the two conditioning streams of the previous layer's latent representations $\pmb{z}^{l-1}$ and the feature map of the ResNet encoder and handle their unbalanced sizes, we design a \textit{residual variational layer}, \elancomment{detailed in the supplementary material.}} The \elancomment{layers'} evidence encoders are independent and consist of two ResNet blocks with feature maps [16, 32], with downsampling applied to each. The decoder consists of two ResNet blocks with transposed convolutions and similar feature maps. We used RELU non-linearties. The model was trained for 1000 epochs with linear deterministic warm-up. The rest of the experimental setup is the same as that of the previous section. As shown in Table \ref{tab:resnet_mnist_marginal}, our model performs better than all recent models that do not use expensive coupling or autoregressive layers and on par with models of higher complexity. The generative and reconstructive capacity of the model is illustrated in Figures \ref{fig:mnist_generated} and \ref{fig:mnist_true}, \ref{fig:mnist_reconstructed}, respectively.
\begin{table}[ht]
\vspace{-0.15in}
\renewcommand{\arraystretch}{0.7}
\centering
\caption{Dynamically binarized MNIST performance for VAEs with sophisticated layers. 1000 importance samples were used for the estimation of the marginal likelihood. All performances listed here are taken from \cite{maaloe2019biva} \ificomment{ and \cite{durkan2019neural}}. All models were trained with a single importance sample.}
\resizebox{0.7\textwidth}{!}{
\begin{tabular}[t]{lcc}
\toprule
Model  &$log \ p(x) \ge $\\
\midrule
\textit{Models with autoregressive (AR) \ificomment{ or coupling (C)} components} & \\ 
VLAE \final{\cite{chen2016variational}} & $-79.03$ \\
Pixel RNN \cite{oord2016pixel} & $-79.20$ \\
\ificomment{RQ-NSF (C)} \cite{durkan2019neural} & $-79.63$ \\
Pixel VAE \cite{gulrajani2016pixelvae} & $-79.66$ \\
\ificomment{RQ-NSF (AR)} \cite{durkan2019neural} & $-79.71$ \\
IAF VAE \cite{kingma2016improved} & $-79.88$ \\
DRAW \cite{gregor2015draw} & $-80.97$\\
Pixel CNN \cite{van2016conditional} & $-81.30$\\
\midrule
\textit{Models without autoregressive \ificomment{or coupling components}} & \\ 
\textbf{SeRe-VAE}  & $\pmb{\final{-79.97 \pm 0.11}}$\\
BIVA \cite{maaloe2019biva} & $-80.47$\\
Discrete VAE \cite{rolfe2016discrete} & $-81.01$  \\
\bottomrule
\label{tab:resnet_mnist_marginal}
\end{tabular}}
\end{table}%
\begin{figure*}[h]
\centering
 \resizebox{0.8\textwidth}{!}{
\begin{subfigure}{0.3\textwidth}
\centering
\makebox[\textwidth][l]{\parbox{0.315\textwidth}{%
  \begin{subfigure}{.3\textwidth}
   \centering
      \includegraphics[width=\linewidth]{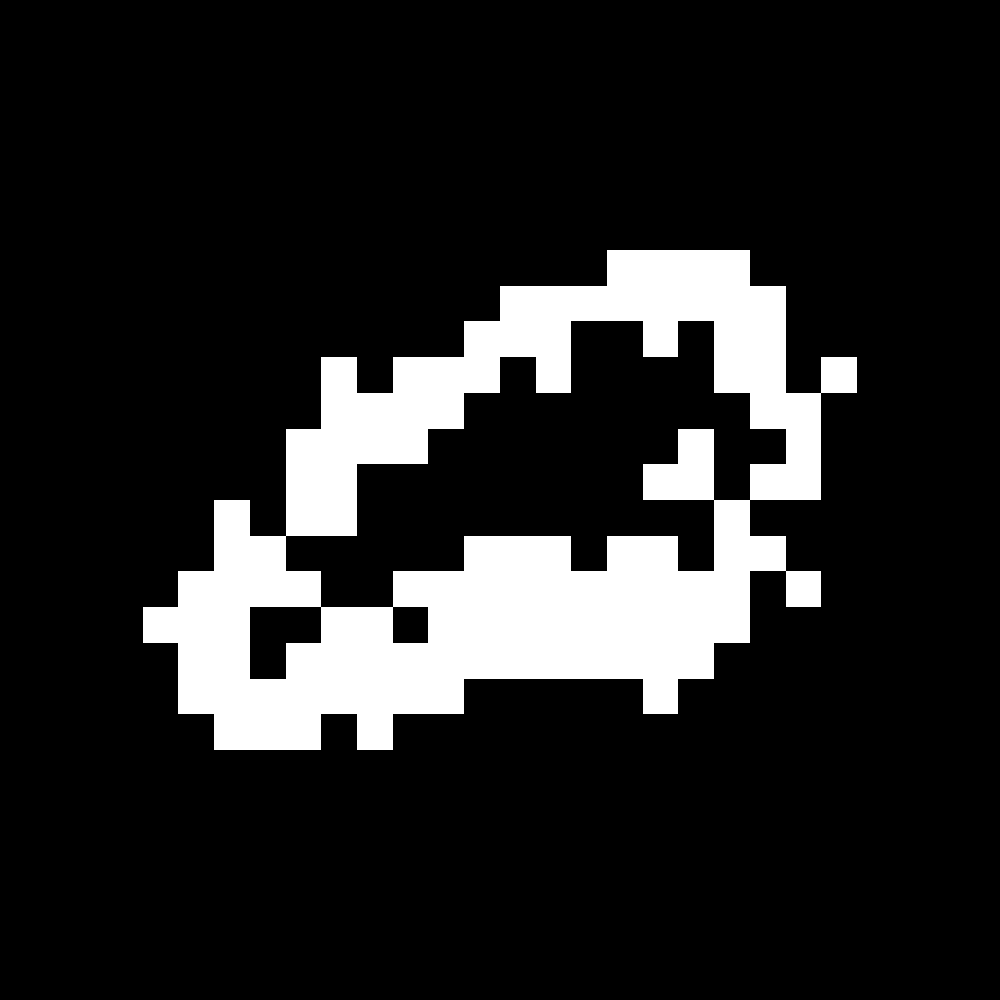}
  \end{subfigure}%
  \begin{subfigure}{.3\textwidth}
   \centering
      \includegraphics[width=\linewidth]{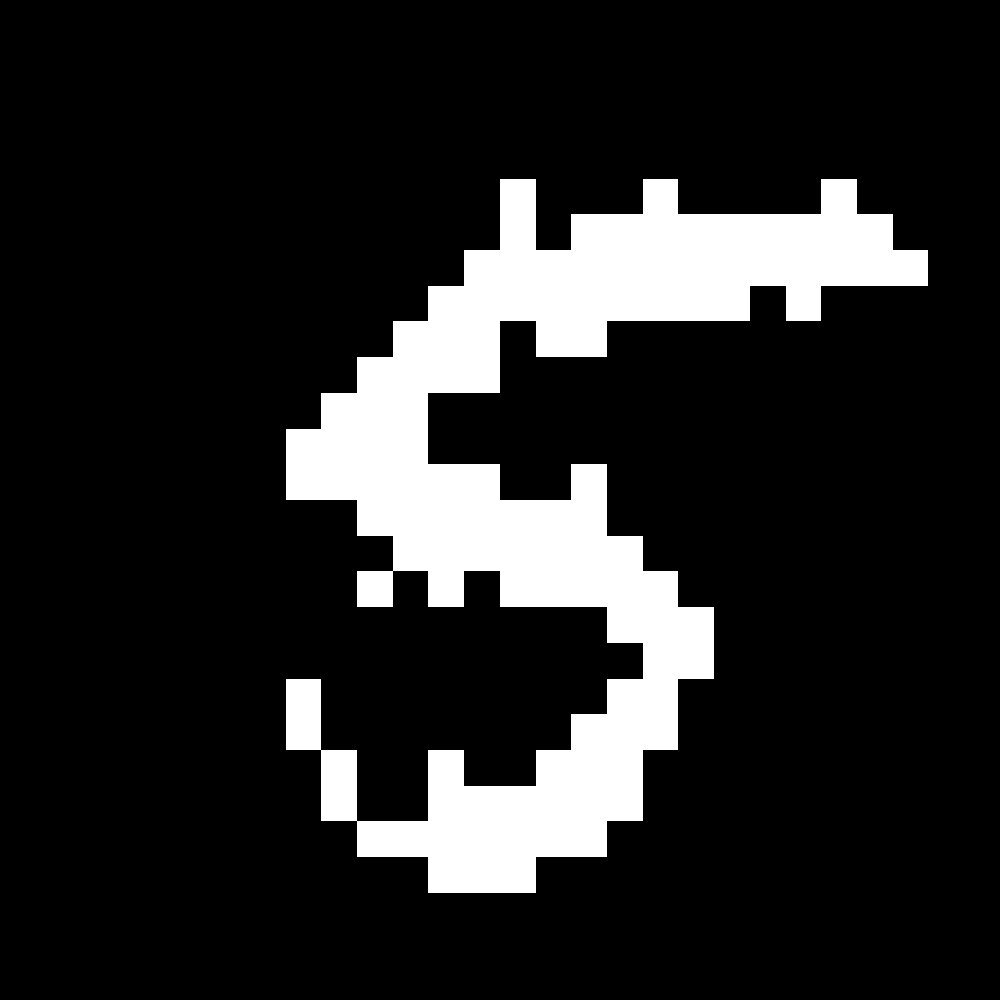}
  \end{subfigure}%
  \begin{subfigure}{.3\textwidth}
   \centering
      \includegraphics[width=\linewidth]{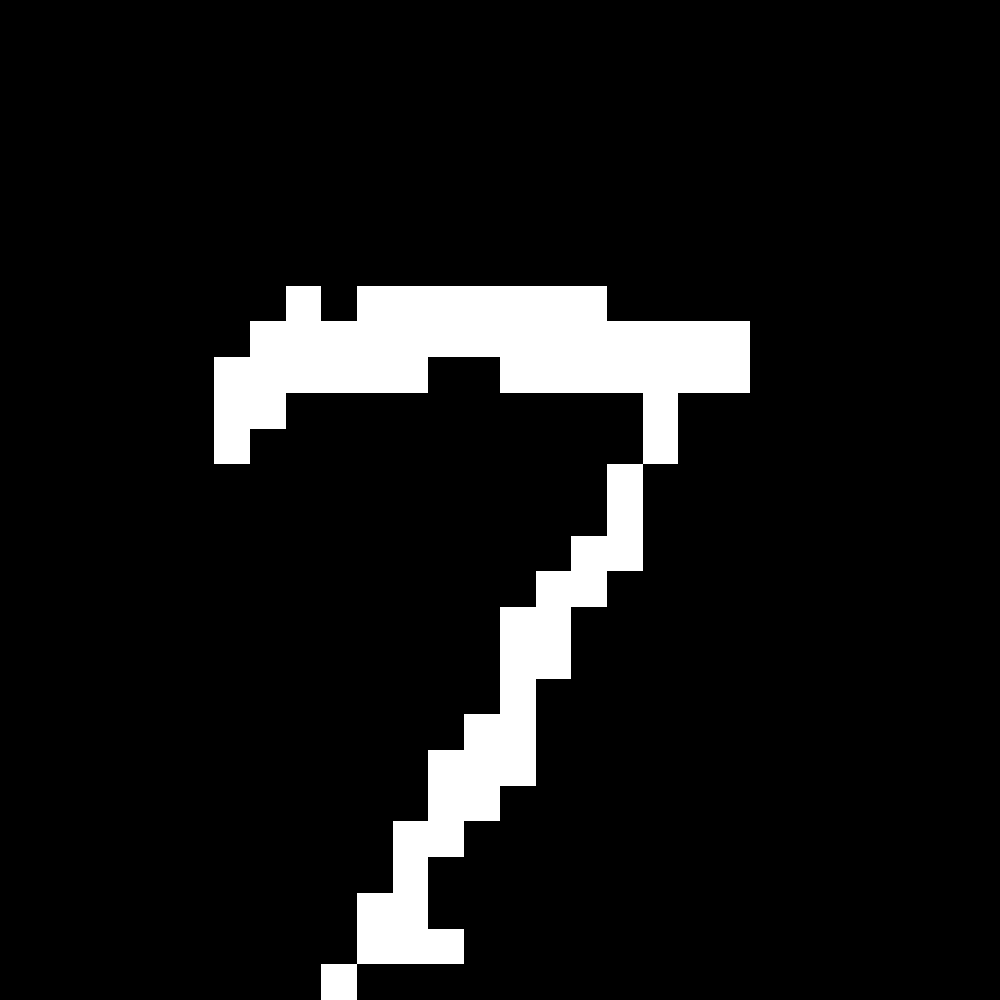}
  \end{subfigure}
    \begin{subfigure}{.3\textwidth}
     \centering
      \includegraphics[width=\linewidth]{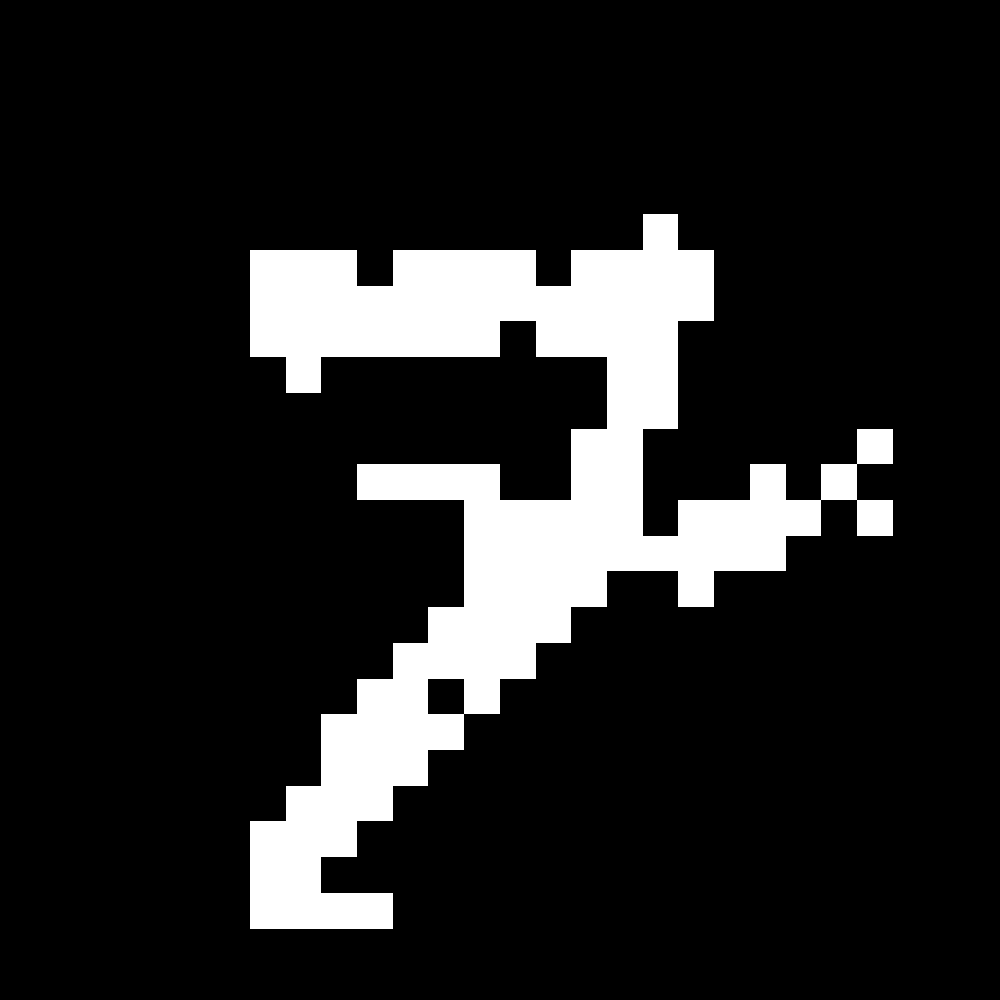}
  \end{subfigure}%
  \begin{subfigure}{.3\textwidth}
   \centering
      \includegraphics[width=\linewidth]{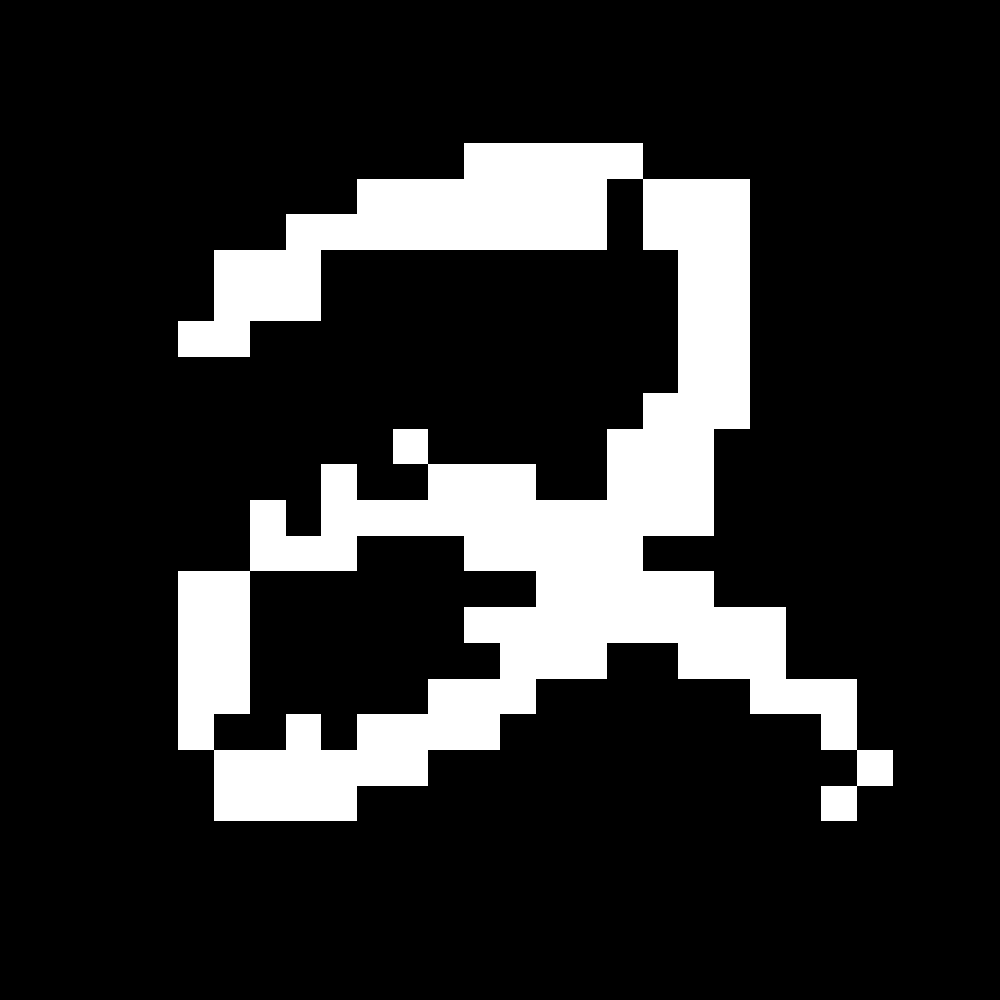}
  \end{subfigure}%
  \begin{subfigure}{.3\textwidth}
      \includegraphics[width=\linewidth]{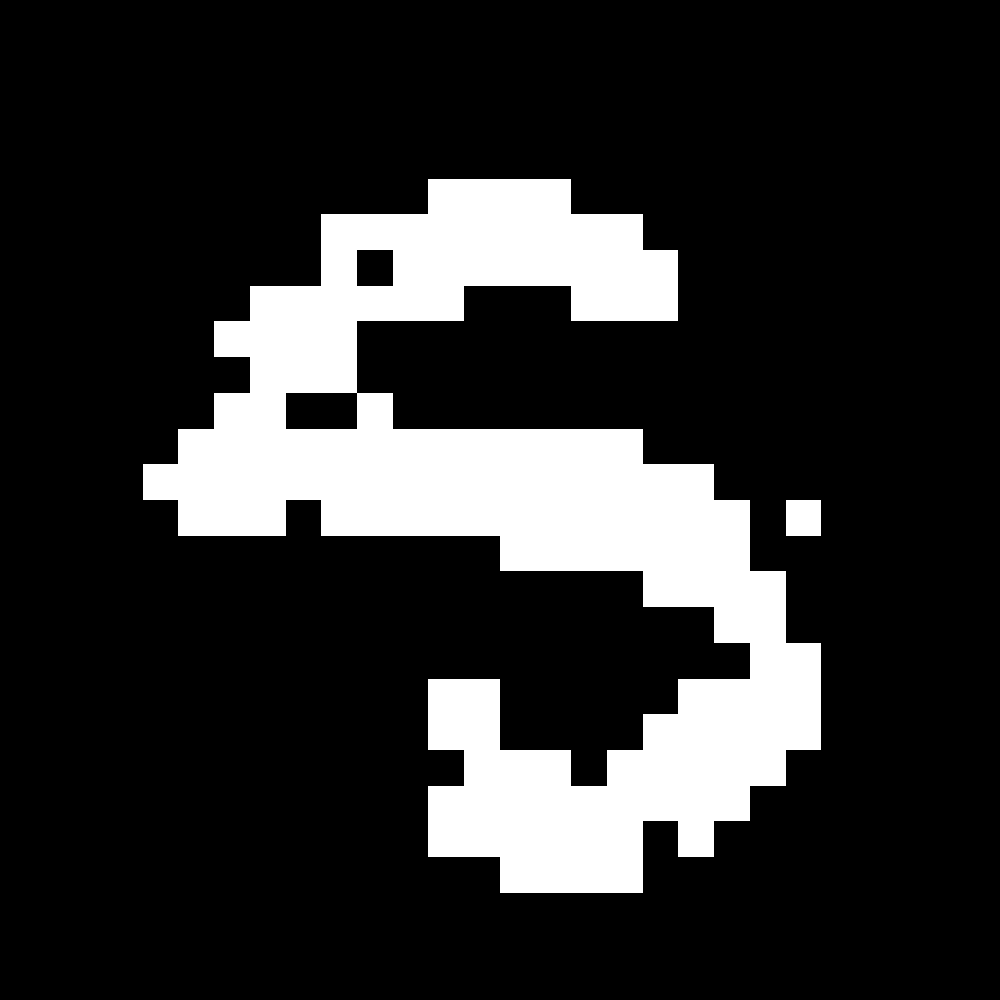}
  \end{subfigure}
    \begin{subfigure}{.3\textwidth}
     \centering
      \includegraphics[width=\linewidth]{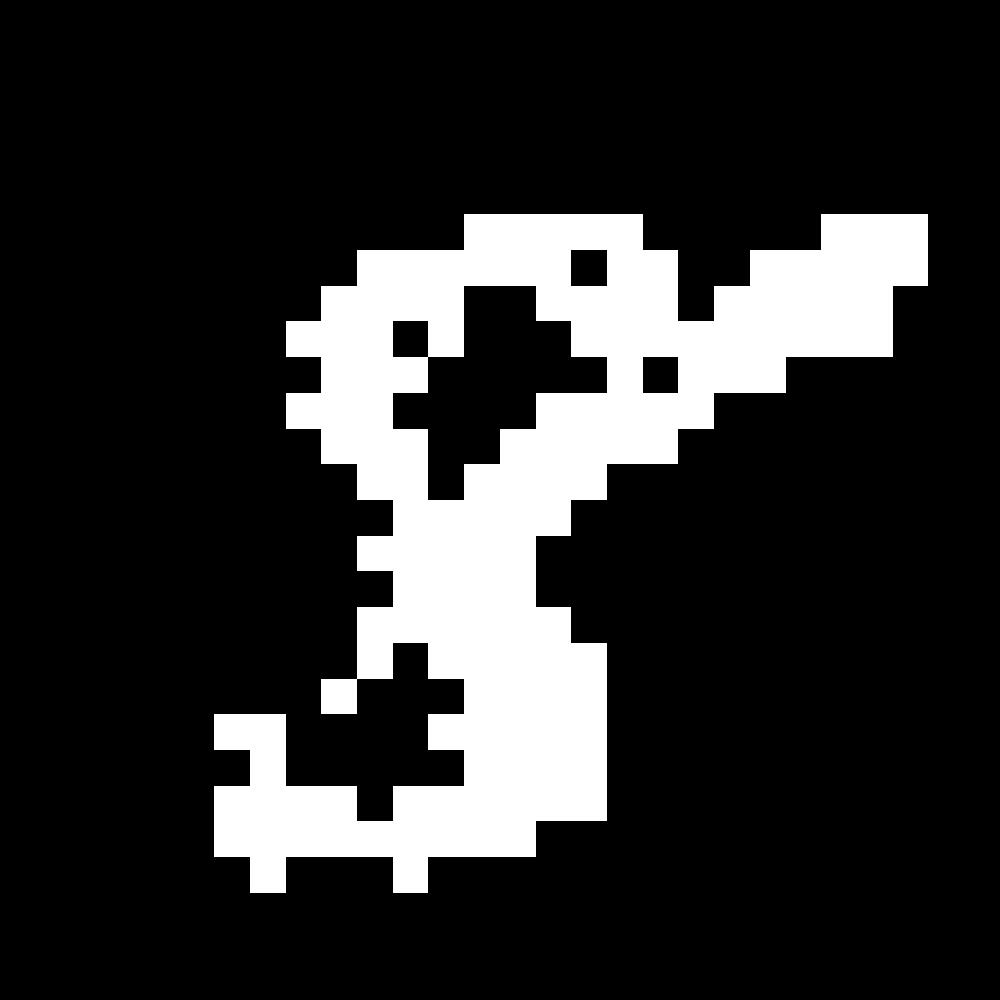}
  \end{subfigure}%
  \begin{subfigure}{.3\textwidth}
   \centering
      \includegraphics[width=\linewidth]{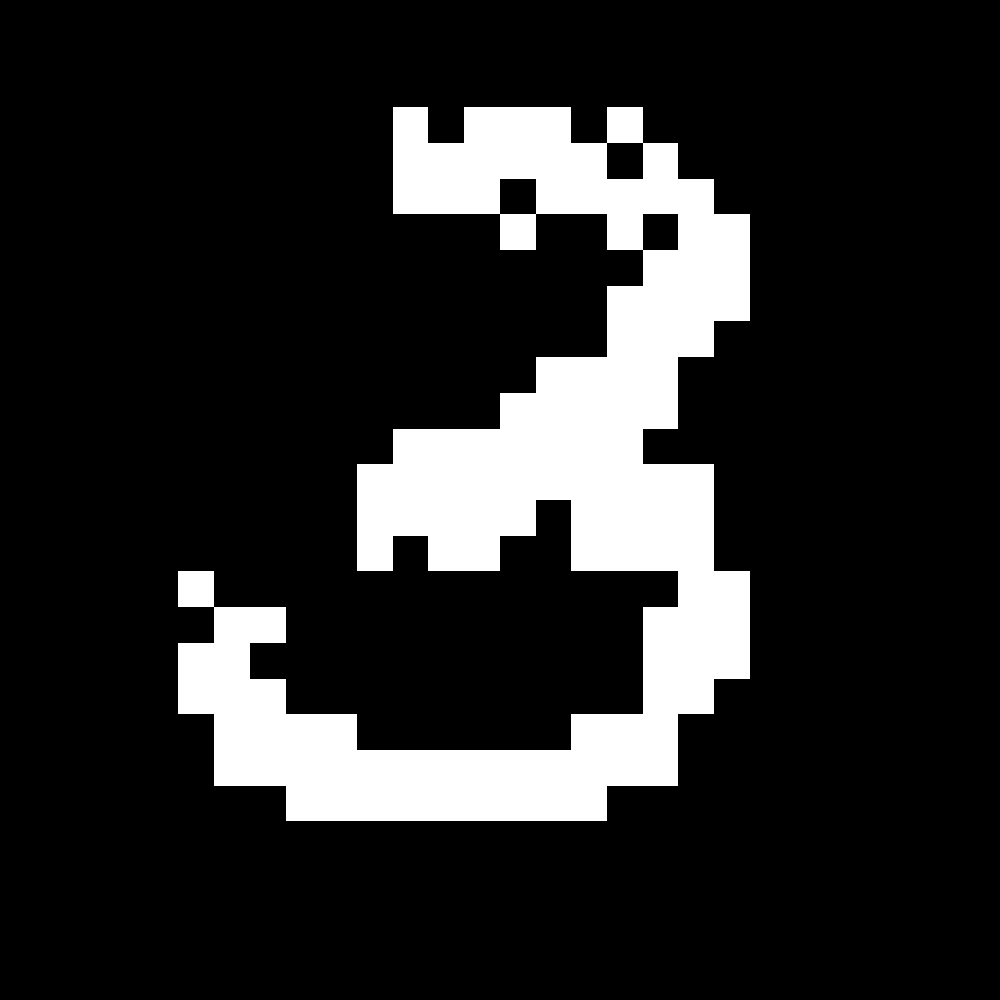}
  \end{subfigure}%
  \begin{subfigure}{.3\textwidth}
   \centering
      \includegraphics[width=\linewidth]{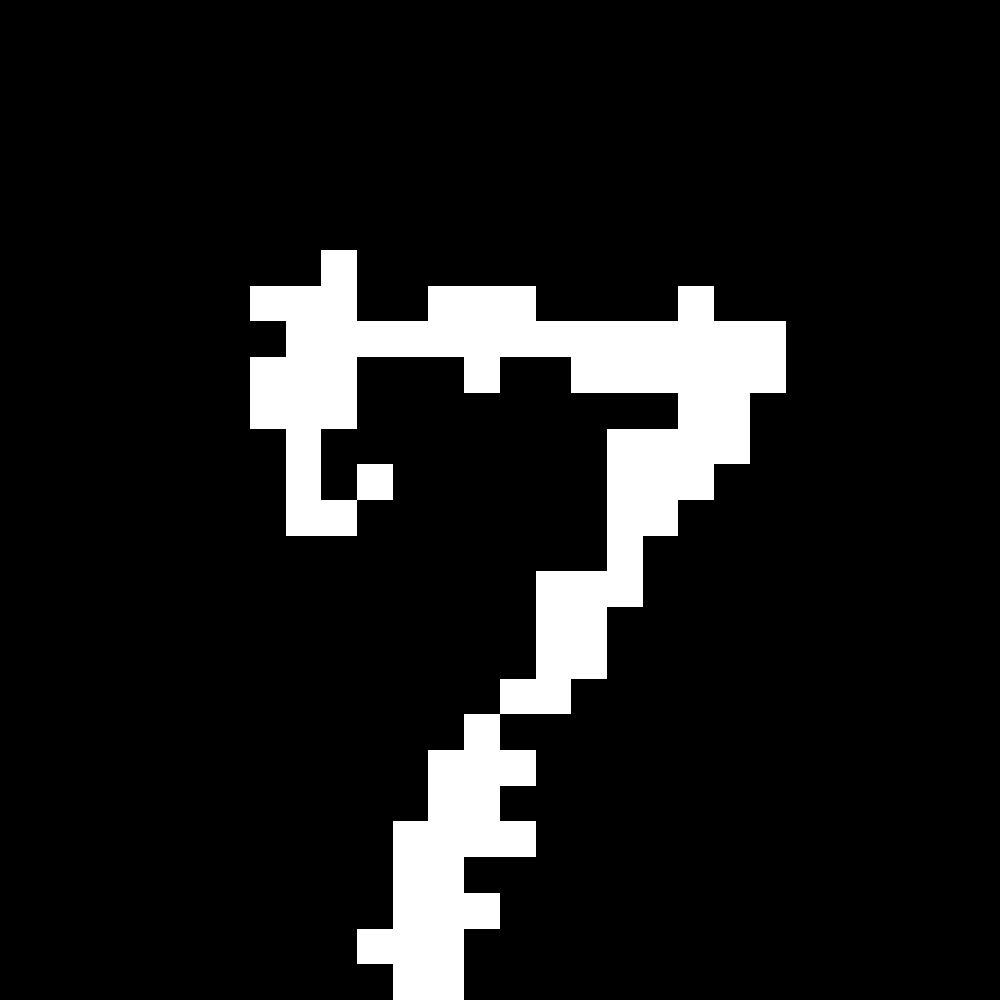}
  \end{subfigure}
}}
 \raggedleft
\caption{Generated MNIST digits}\label{fig:mnist_generated}
\end{subfigure}
\hspace{0.12in}
 \begin{subfigure}{0.3\textwidth}
\makebox[\textwidth][l]{\parbox{0.3\textwidth}{%
  \begin{subfigure}{.3\textwidth}
      \includegraphics[width=\linewidth]{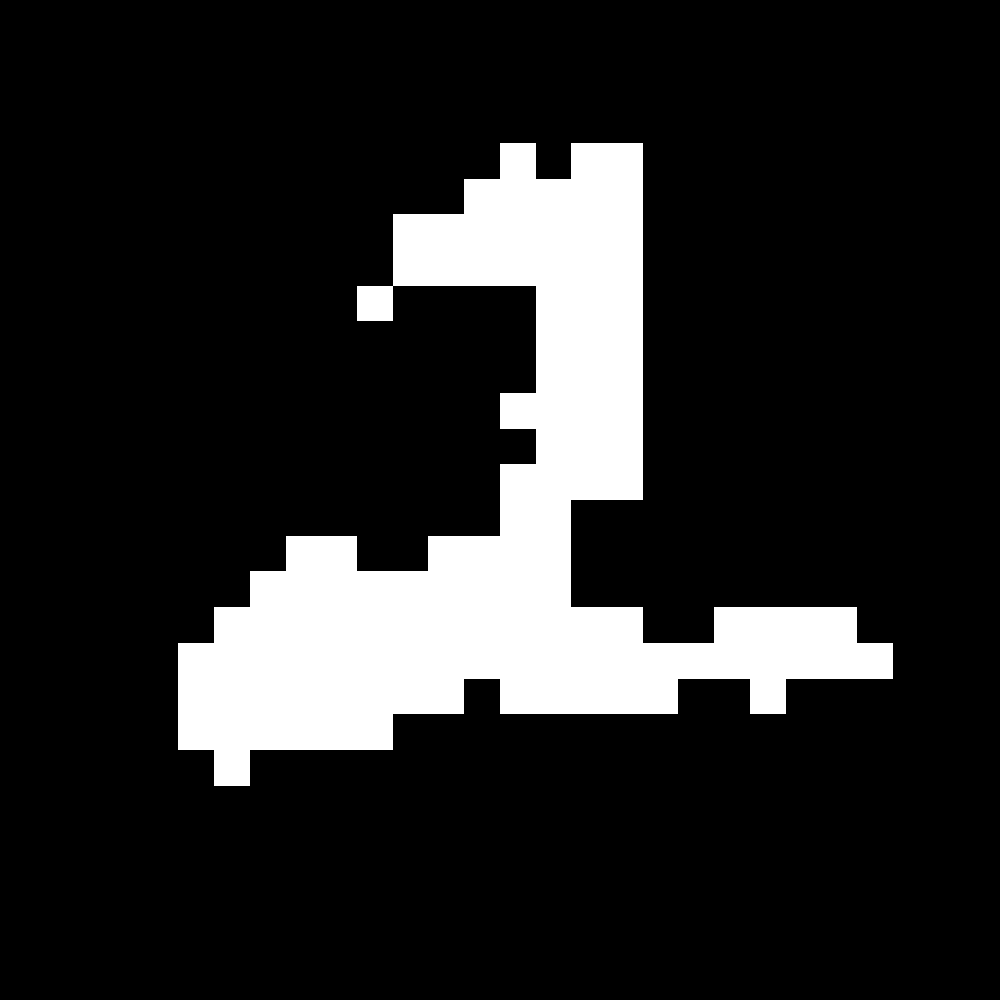}
  \end{subfigure}%
  \begin{subfigure}{.3\textwidth}
      \includegraphics[width=\linewidth]{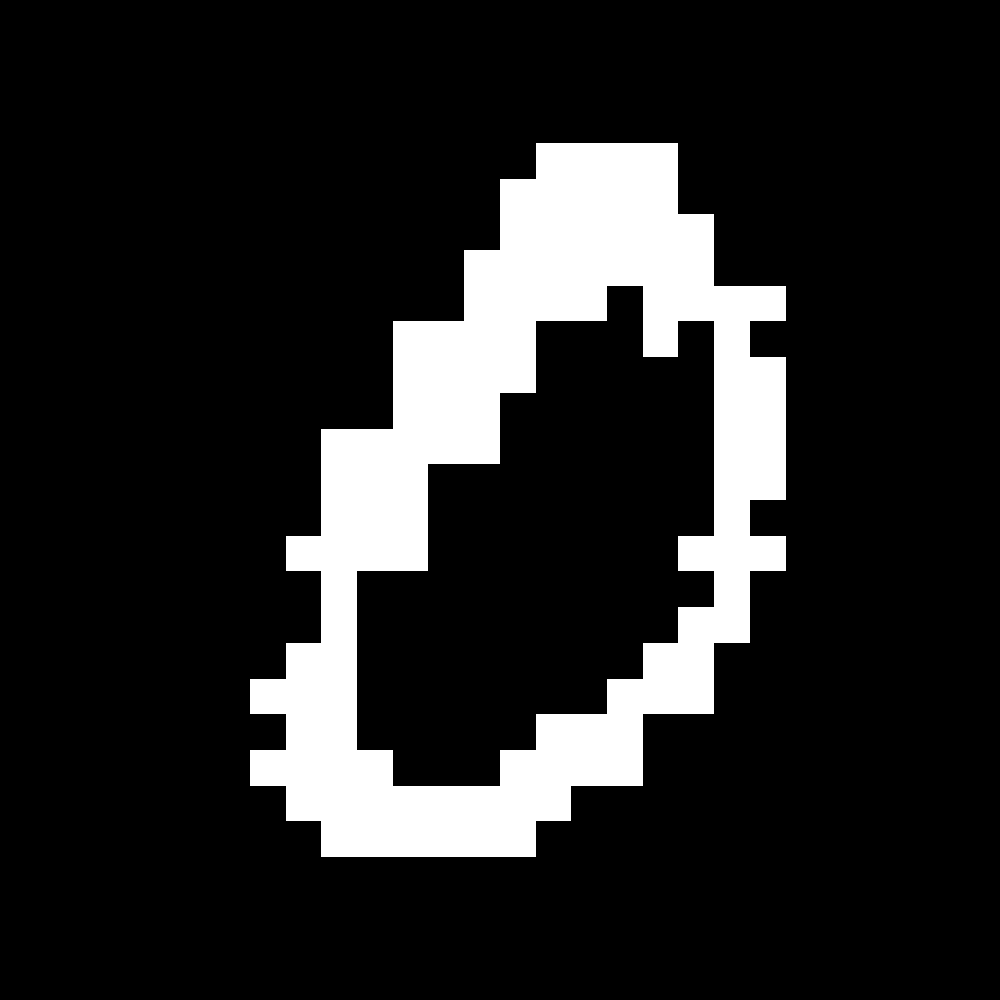}
  \end{subfigure}%
  \begin{subfigure}{.3\textwidth}
      \includegraphics[width=\linewidth]{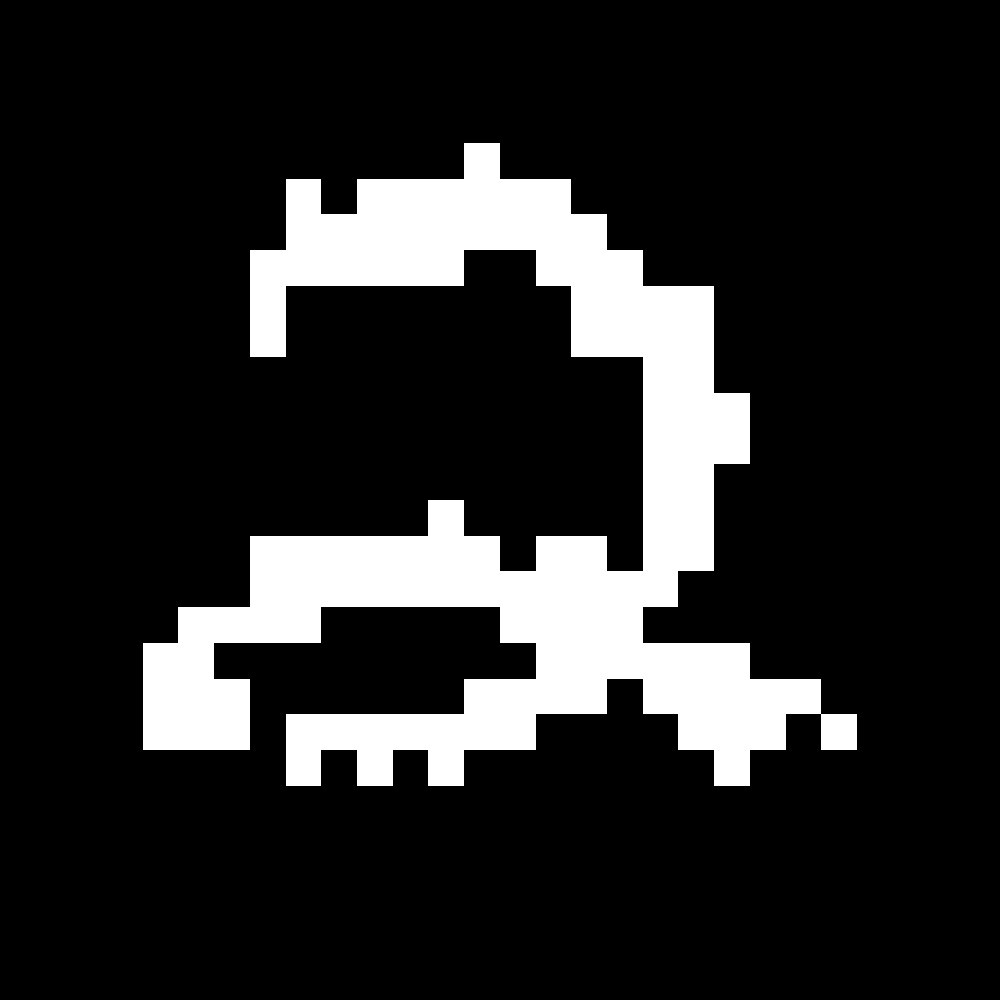}
  \end{subfigure}
    \begin{subfigure}{.3\textwidth}
      \includegraphics[width=\linewidth]{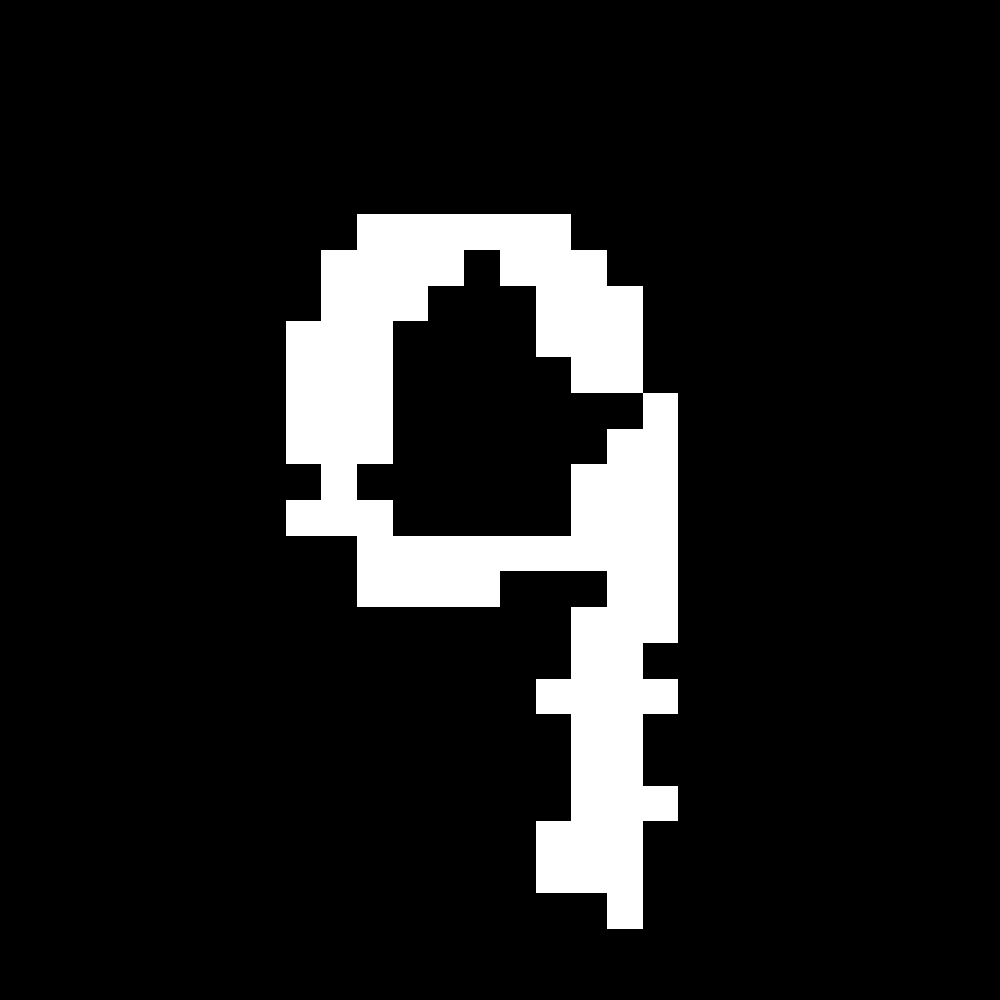}
  \end{subfigure}%
  \begin{subfigure}{.3\textwidth}
      \includegraphics[width=\linewidth]{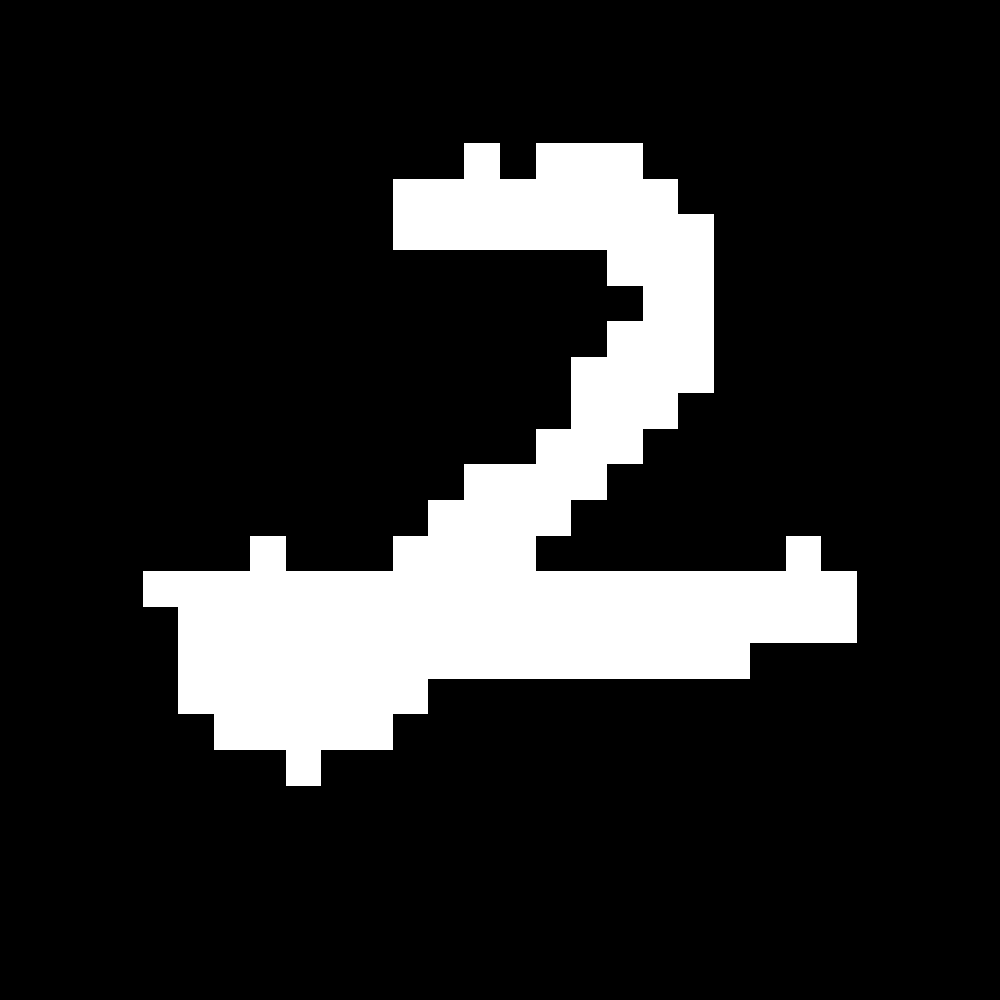}
  \end{subfigure}%
  \begin{subfigure}{.3\textwidth}
      \includegraphics[width=\linewidth]{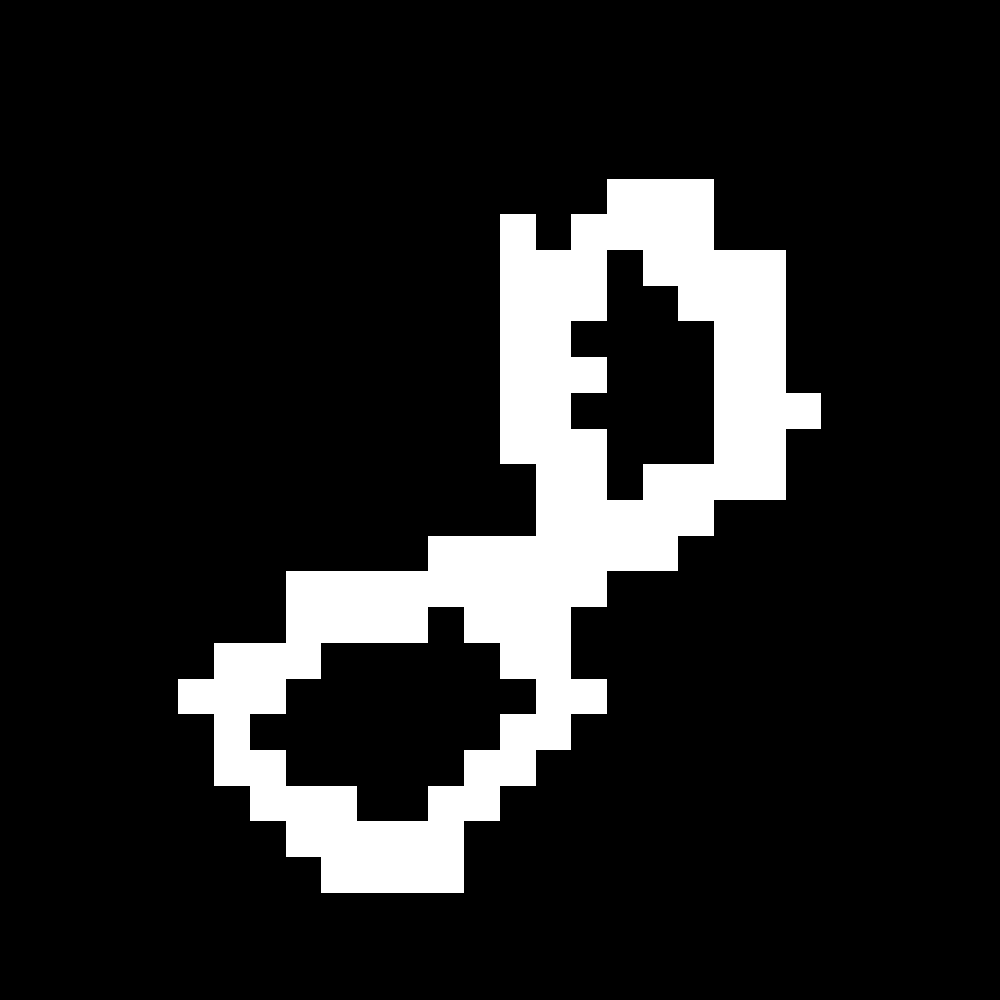}
  \end{subfigure}
    \begin{subfigure}{.3\textwidth}
      \includegraphics[width=\linewidth]{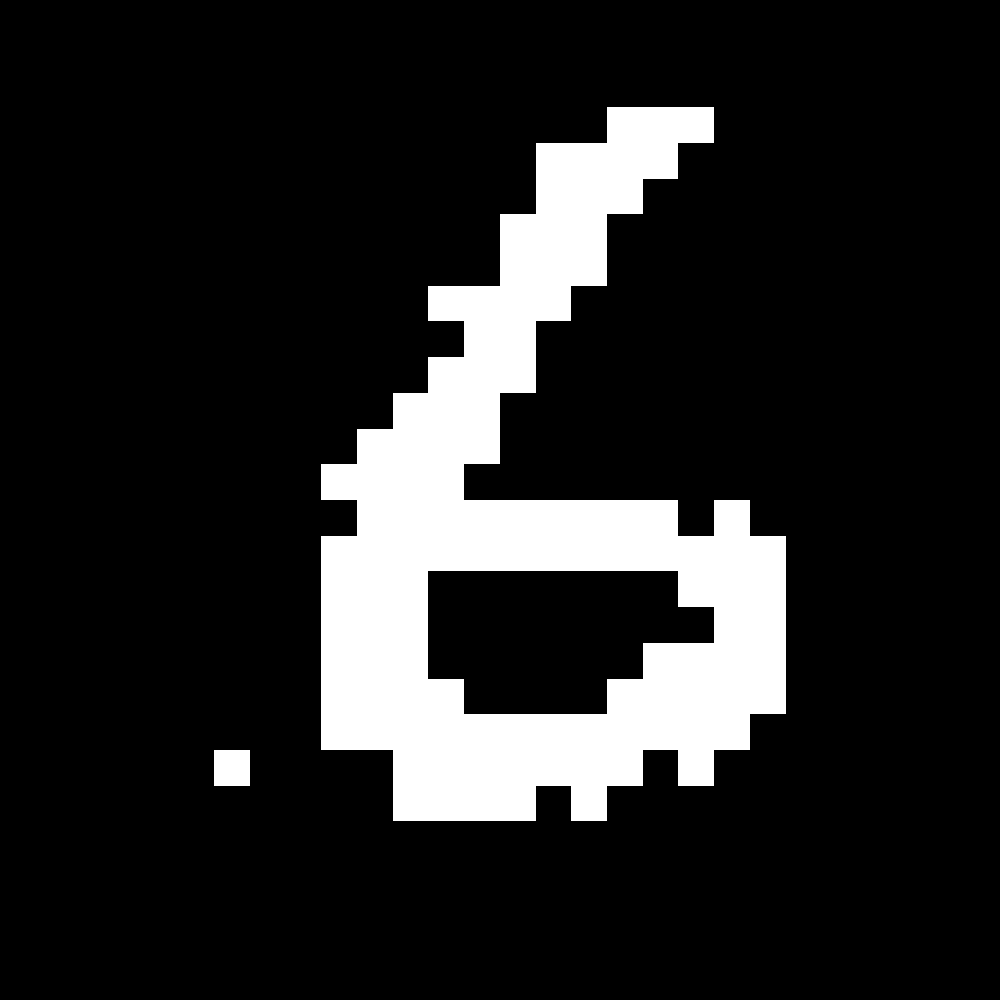}
  \end{subfigure}%
  \begin{subfigure}{.3\textwidth}
      \includegraphics[width=\linewidth]{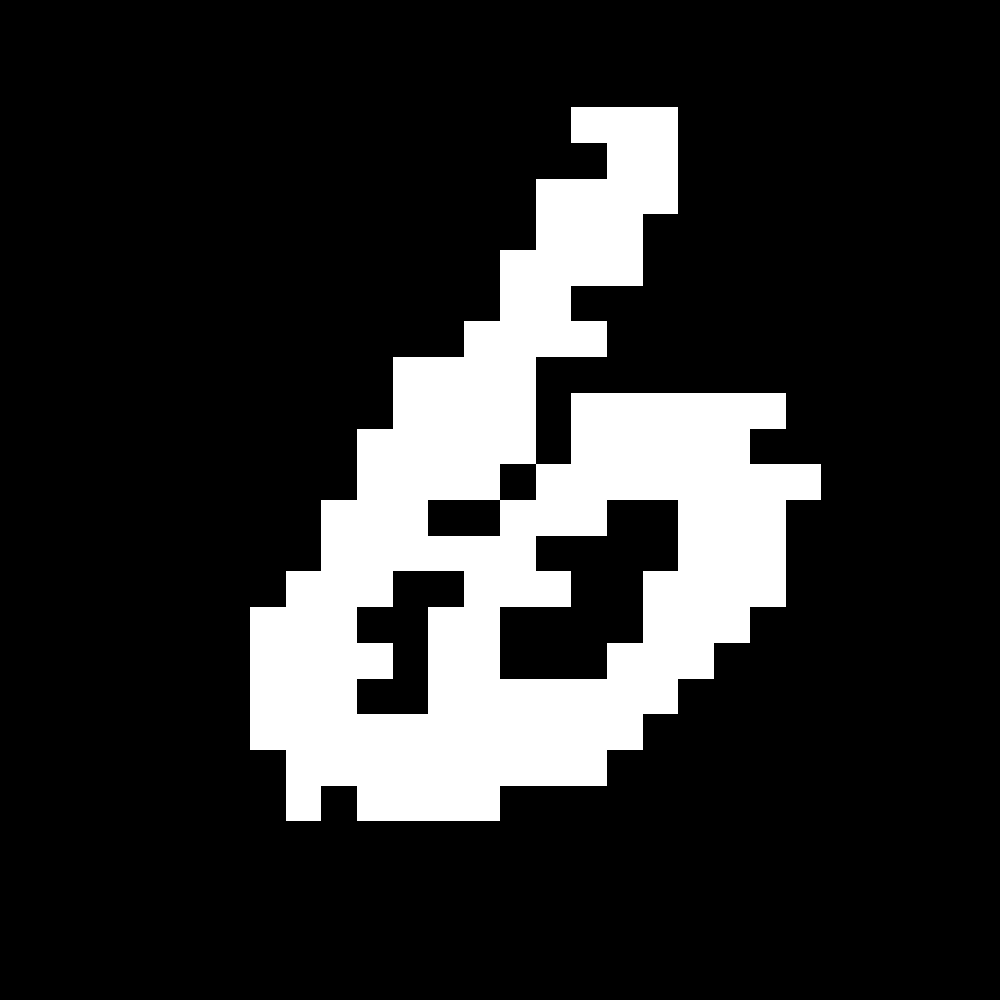}
  \end{subfigure}%
  \begin{subfigure}{.3\textwidth}
      \includegraphics[width=\linewidth]{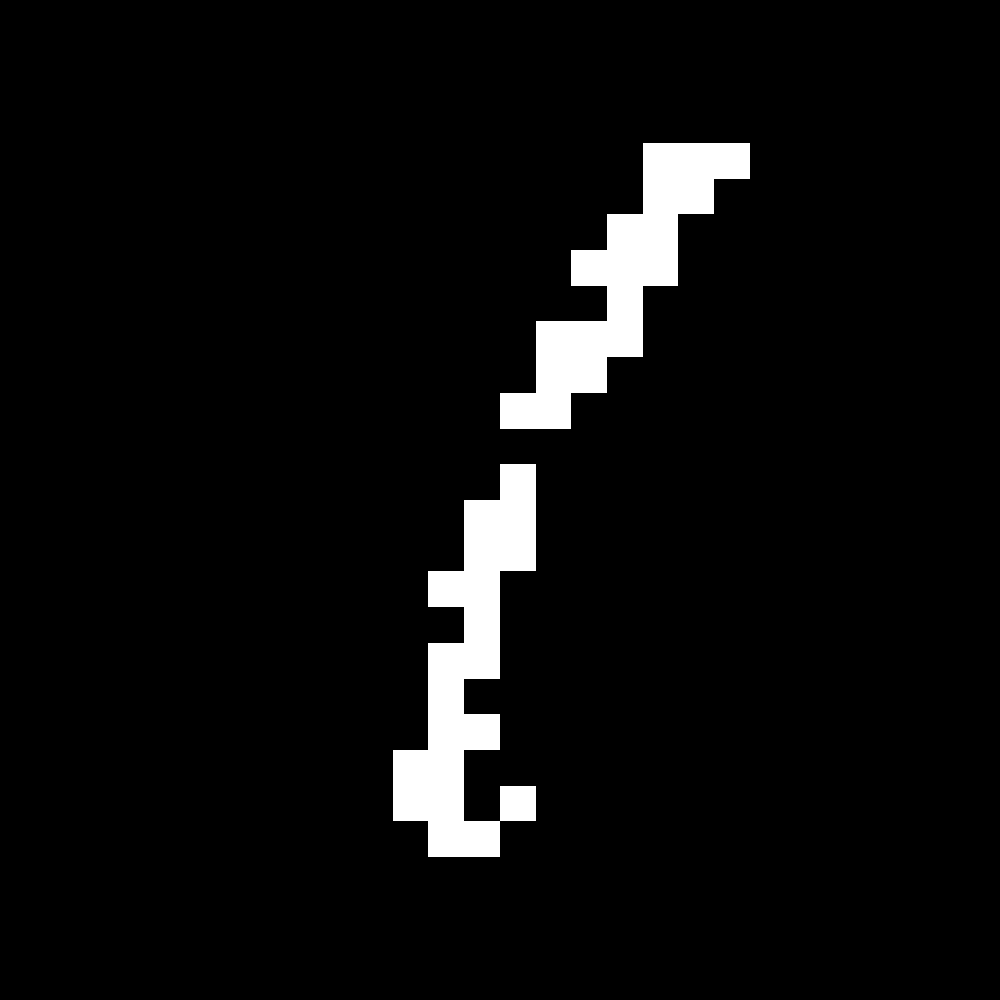}
  \end{subfigure}
}
}
 \raggedleft
\caption{True MNIST digits}
\label{fig:mnist_true}
 \end{subfigure}
\hspace{0.12in}
 \begin{subfigure}{0.3\textwidth}
 \centering
\makebox[\textwidth][l]{\parbox{0.3\textwidth}{%
  \begin{subfigure}{.3\textwidth}
   \centering
      \includegraphics[width=\linewidth]{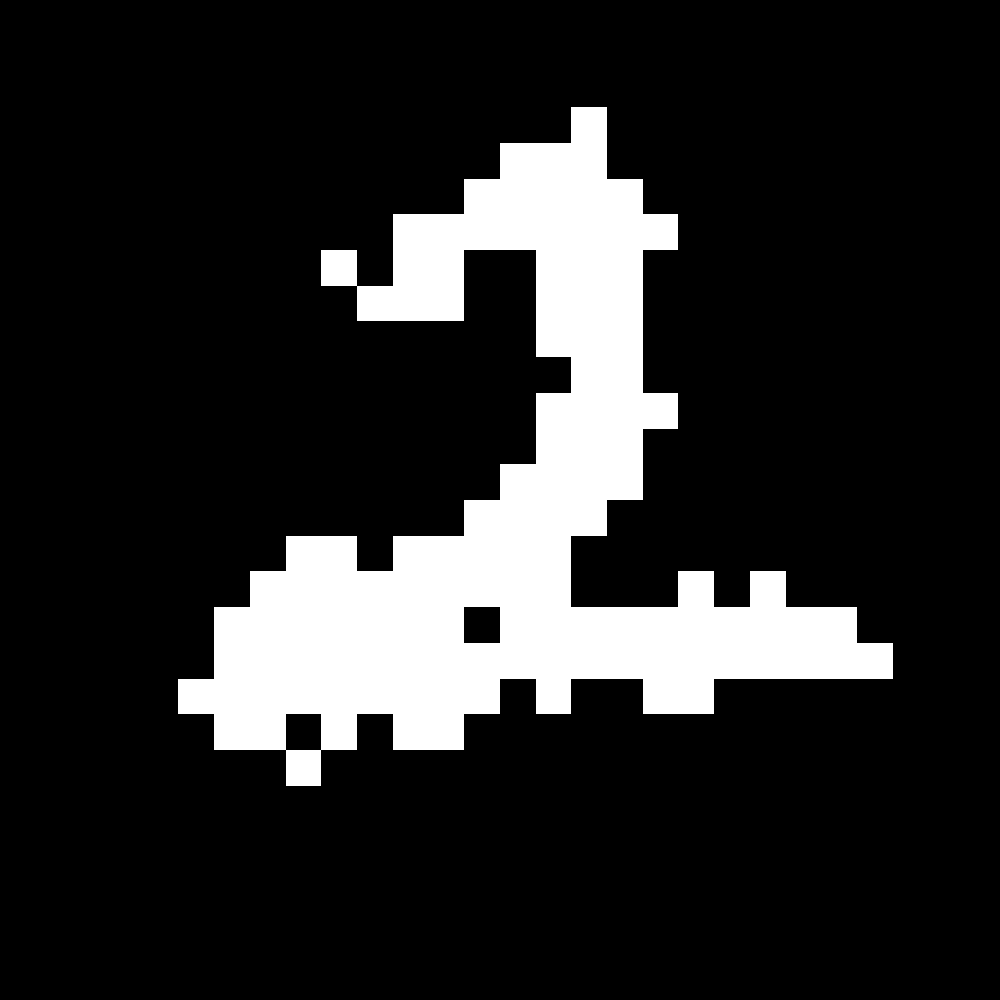}
  \end{subfigure}%
  \begin{subfigure}{.3\textwidth}
   \centering
      \includegraphics[width=\linewidth]{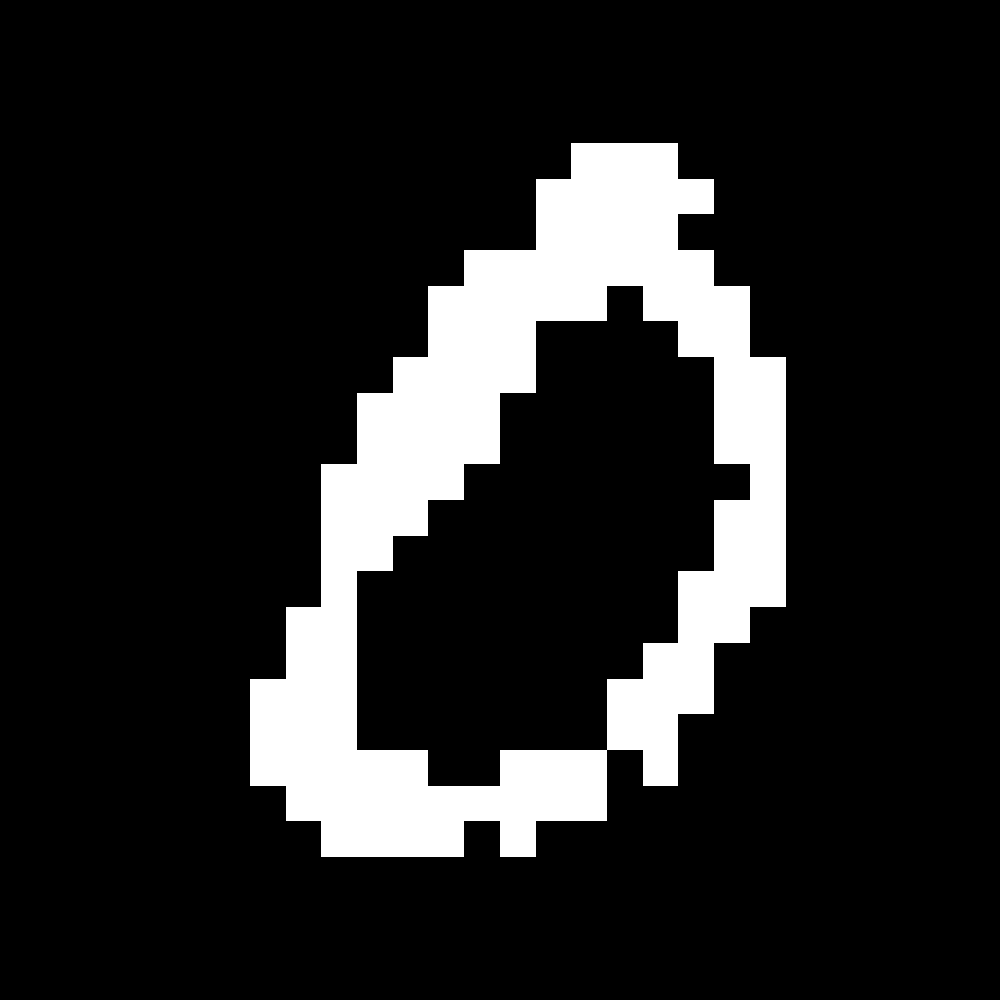}
  \end{subfigure}%
  \begin{subfigure}{.3\textwidth}
   \centering
      \includegraphics[width=\linewidth]{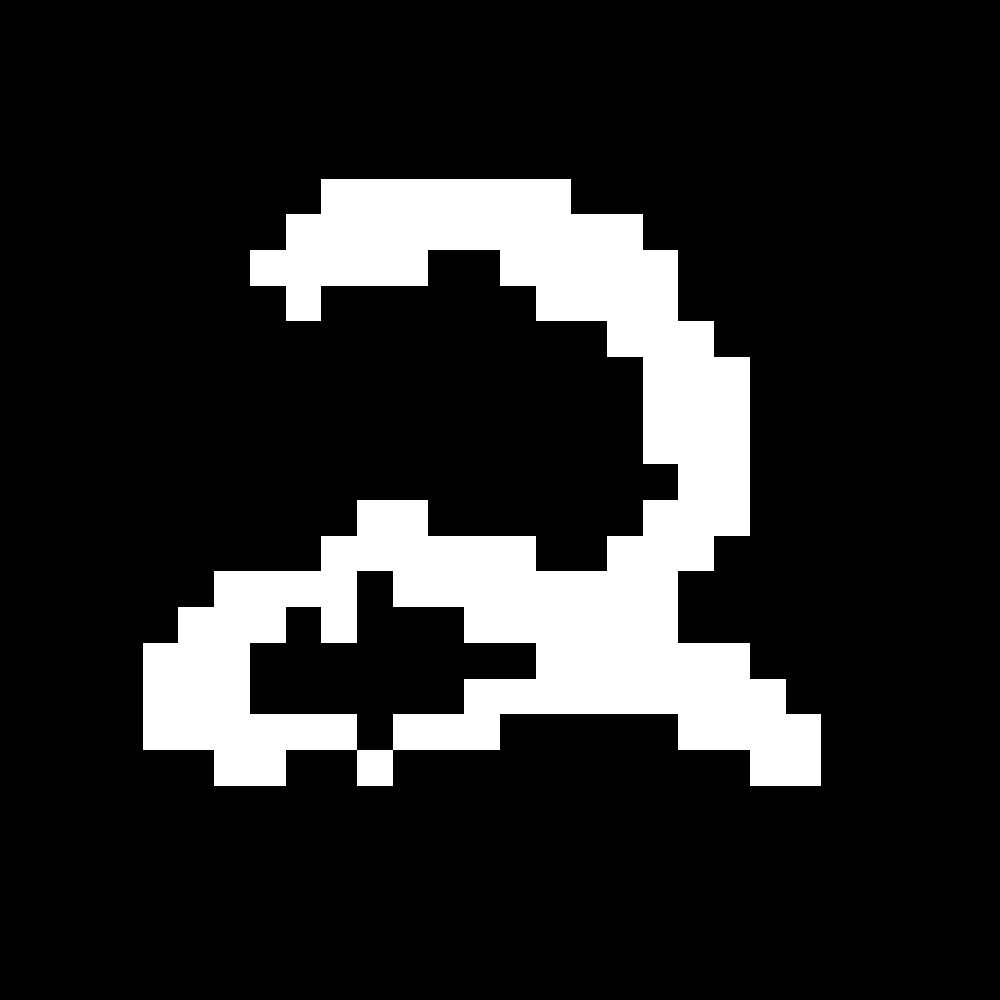}
  \end{subfigure}
    \begin{subfigure}{.3\textwidth}
      \includegraphics[width=\linewidth]{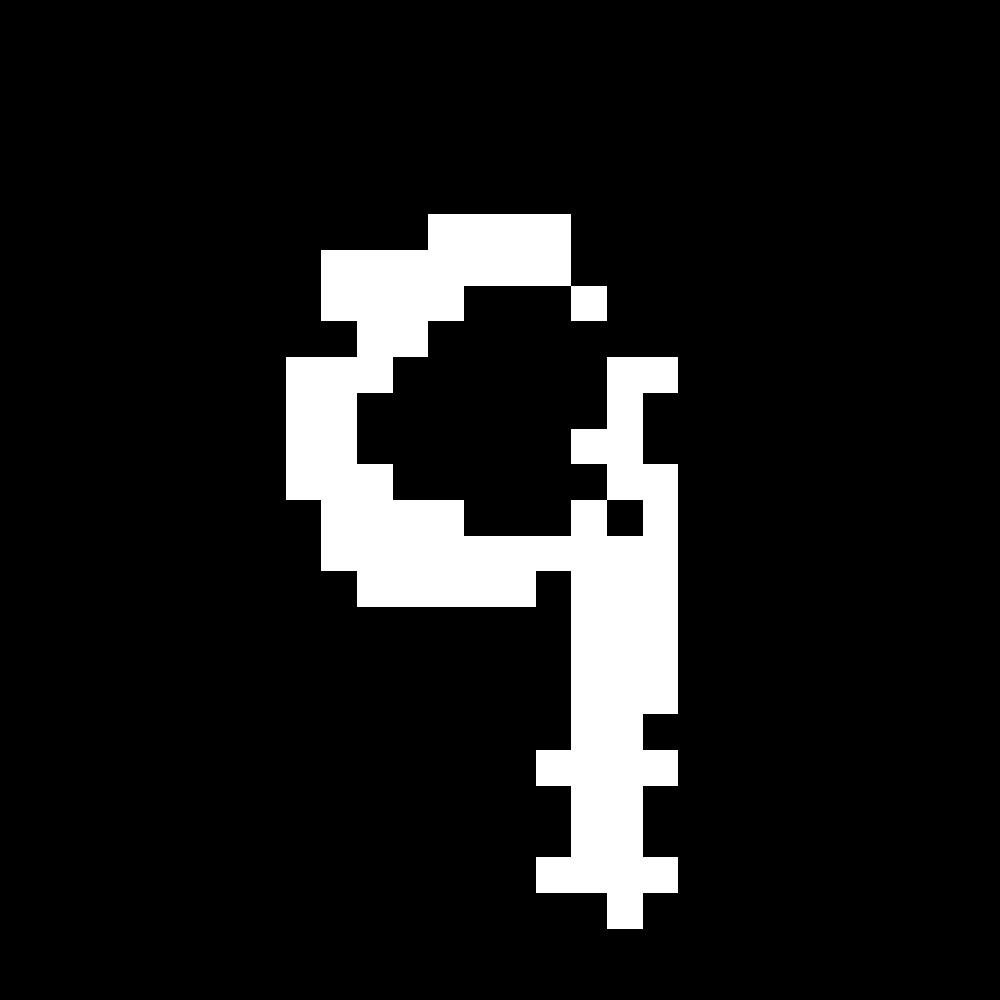}
  \end{subfigure}%
  \begin{subfigure}{.3\textwidth}
      \includegraphics[width=\linewidth]{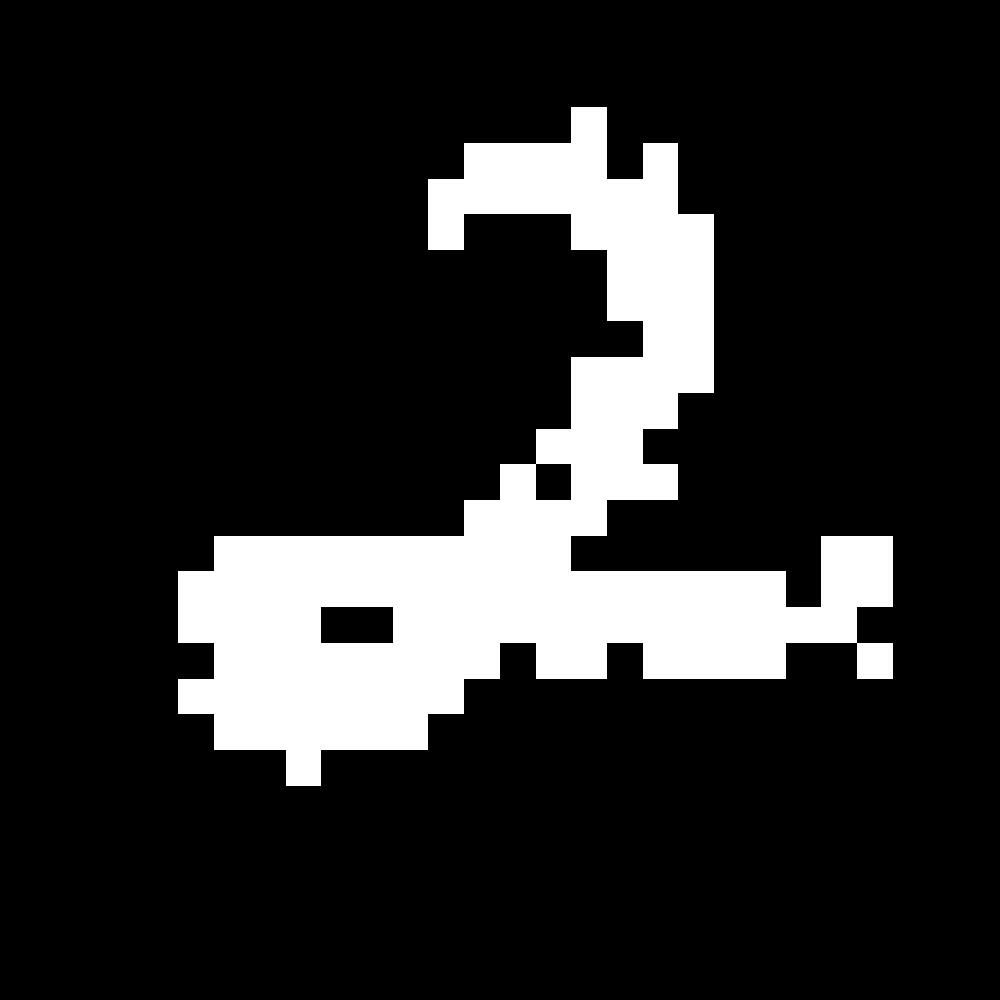}
  \end{subfigure}%
  \begin{subfigure}{.3\textwidth}
   \centering
      \includegraphics[width=\linewidth]{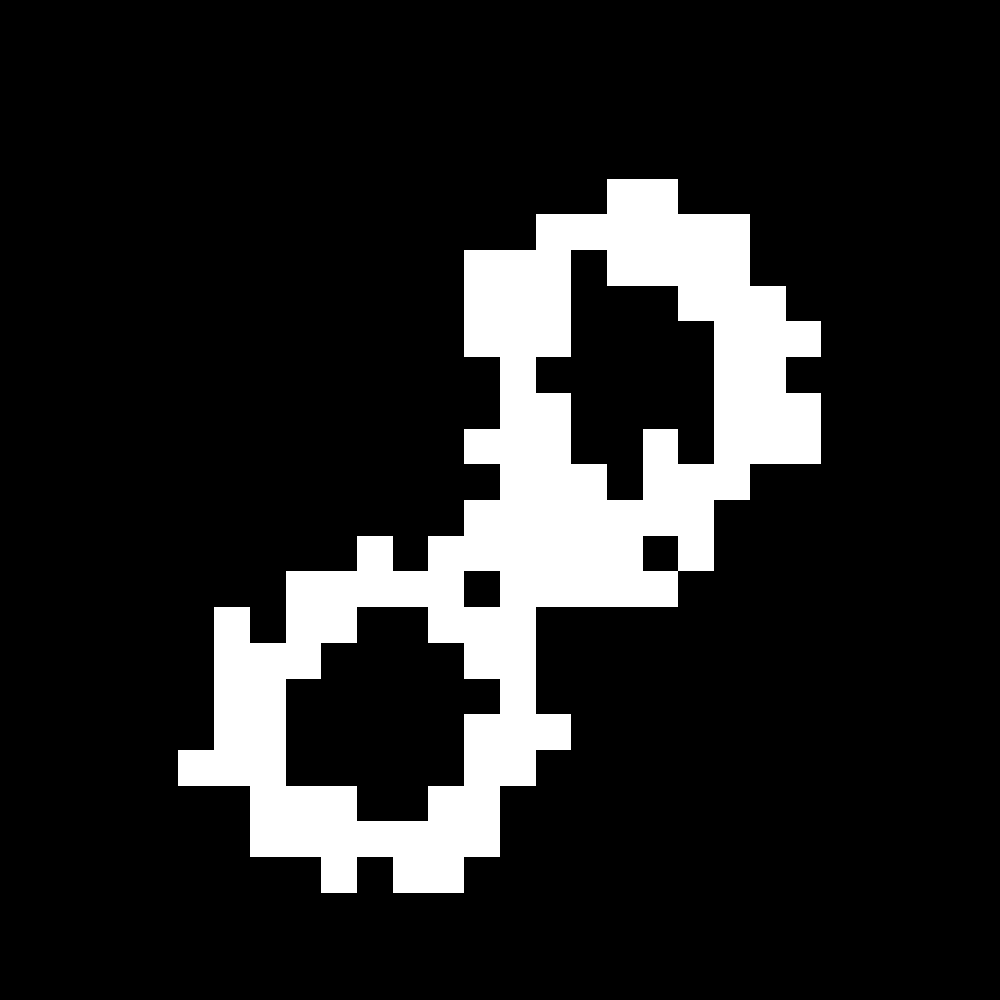}
  \end{subfigure}
    \begin{subfigure}{.3\textwidth}
     \centering
      \includegraphics[width=\linewidth]{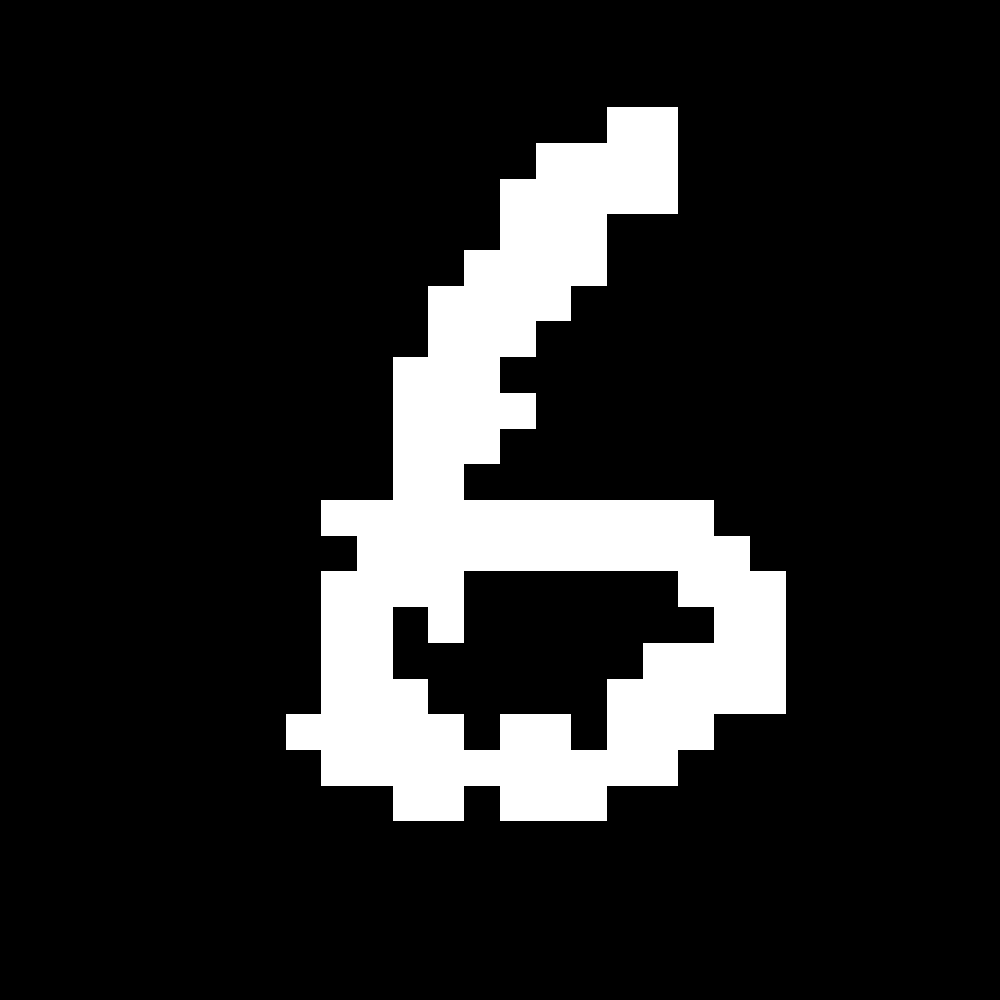}
  \end{subfigure}%
  \begin{subfigure}{.3\textwidth}
   \centering
      \includegraphics[width=\linewidth]{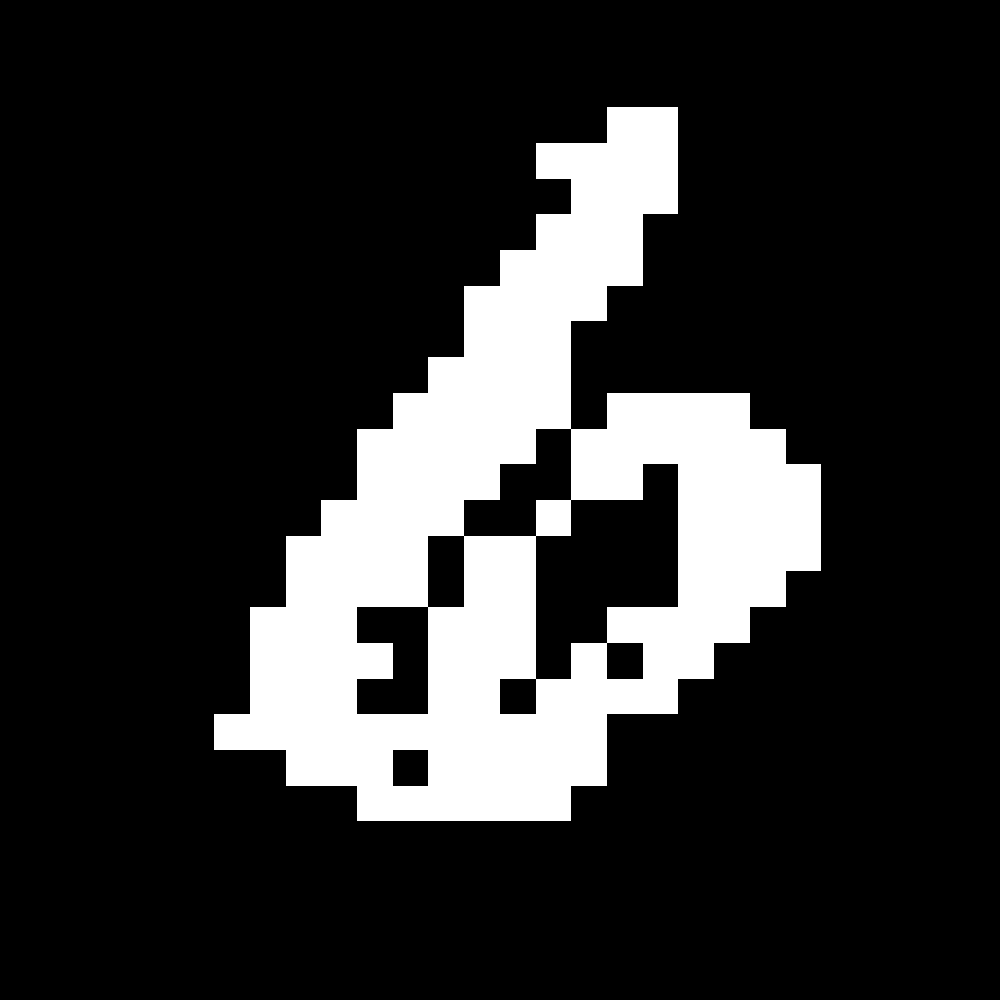}
  \end{subfigure}%
  \begin{subfigure}{.3\textwidth}
   \centering
      \includegraphics[width=\linewidth]{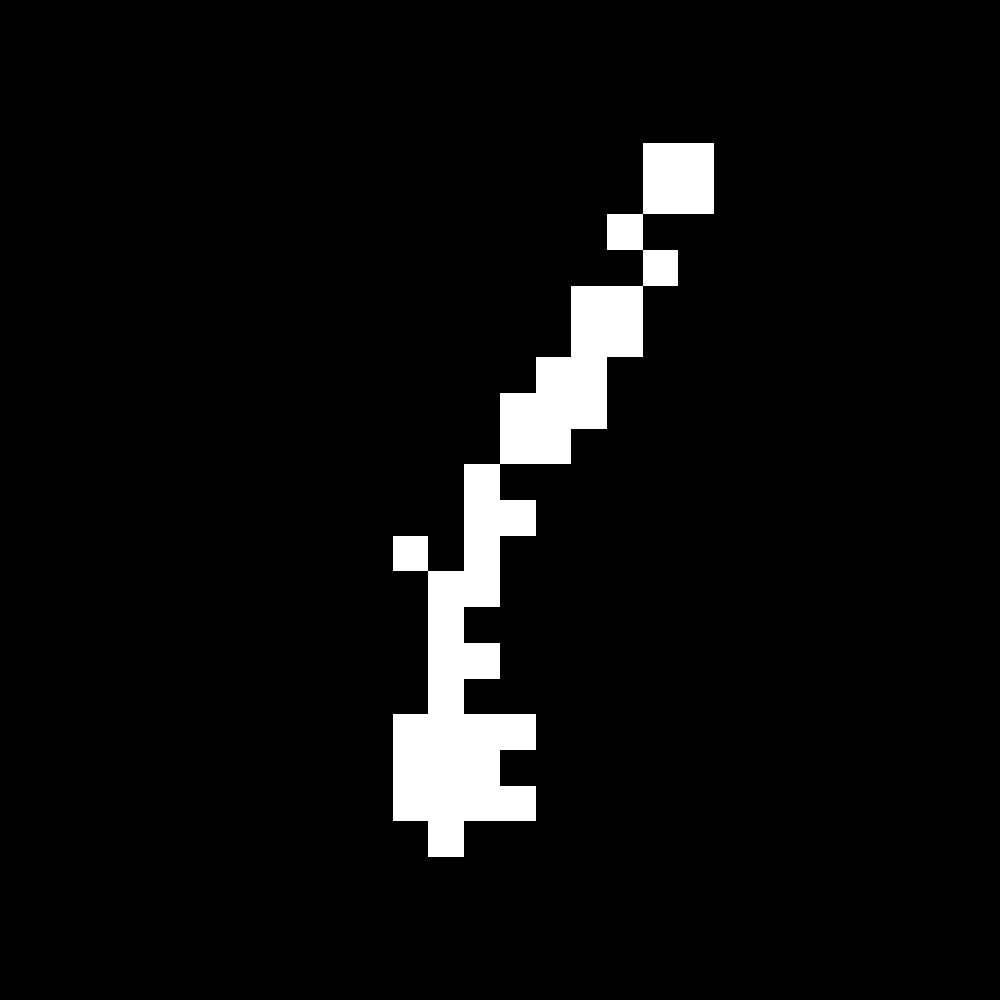}
  \end{subfigure}
}}
 \raggedleft
\caption{Reconstructed MNIST digits}
\label{fig:mnist_reconstructed}
\end{subfigure}}
\caption{Qualitative Performance of the ResNet \ificomment{SeRe-VAE}}
\end{figure*}
\todo{\textbf{TODO: Add nearest neighbors comparisons to the generated images in 7(a).}}
\subsection{Performance of the Self-Reflective Masked Autoregressive Flow on CIFAR-10}\label{sec: self_refl_maf_experiments}
\ificomment{In this section, we validate the variational normalizing flows introduced in Section \ref{sec:maf}. We used a 5 layer hierarchy \corrections{of} $40$ latent variables each and $T=2$ flows. \elancomment{We adopted a \corrections{unit rank} Gaussian base distribution---parameterized as in Equation (9) in \cite{rezende2014stochastic}---and diagonal Gaussian prior and posterior layers.}} We used neural spline generative layers \final{with coupling transformations }\cite{durkan2019neural}, which boosted the performance compared to the affine transformations. The full architectural and training details are provided in the supplementary material. In Table \ref{tab:maf_cifar10}, we compare against generative MAF models with the same or larger width, with or without training dataset augmentation with horizontal image flips and different number of MADEs. \elancomment{Our variational model exhibits significant improvement over the baselines.} 

 \begin{table}[!htbp] \centering
 \caption{ Performance of different MAFs on CIFAR-10.}
 \resizebox{0.95\textwidth}{!}{
 \begin{tabular}{@{\extracolsep{5pt}} cccccc} 
 \\[-1.8ex]\hline 
 \hline \\[-1.8ex] 
 \multicolumn{1}{c}{Model} & \multicolumn{1}{c}{Variational} & \multicolumn{1}{c}{$\#$MADE layers} & \multicolumn{1}{c}{Width} & \multicolumn{1}{c}{Flipped Images} & 
 \multicolumn{1}{c}{Test Loglikelihood} \\
 \hline \\[-1.8ex] 
 \hline{\textbf{SeRe-MAF}}  &  \textbf{Yes}  &  \textbf{10 (2 flows, 5 layers)}  &  \textbf{1024}  &  \textbf{No} &  $\boldsymbol{\ge 3190}$ (\textbf{ELBO}) \\
 \hline{MAF}  &  No  &  10  &  1024  &  No & $ 2670$ \\
 \hline{MAF (5)} \cite{papamakarios2017masked}  &  No  &  5  &  2048  & Yes & 2936   \\
  \hline{MAF (10)} \cite{papamakarios2017masked}  &  No  &  10  &  2048  & Yes & 3049   \\
 \hline \\[-1.8ex]
 \end{tabular}
  \label{tab:maf_cifar10}
  }
 \end{table}
 
\elancomment{We expect further performance improvements could be made by implementing any combination of the following steps: larger MADE layers; more latent features or layers \final{in the hierarchy}; a distribution for the decoder that is specific to natural images such as \cite{oord2016pixel,kingma2018glow}; training with more importance samples \cite{burda2015importance};} \ificomment{and further finetuning of the architectural or training hyperparameters. However, such optimizations are beyond the scope of this paper and are left for future research.}
\section{Conclusion and Discussion}
\ificomment{We presented self-reflective variational inference, a \elancomment{structural modification for hierarchical} VAEs that combines top-down inference and \corrections{iterative feedback} between the generative and inference network \corrections{through shared transformational layers}. 
We also introduced hierarchical latent variable normalizing flows that utilize the proposed architecture to refine the base distribution and the bijectors in a recurrent manner. We used uncoupled deterministic encoders; this facilitates even distribution of the KL\elancomment{-divergence} across the layers and hence meaningful latent representations.}
\ificomment{It would be of interest to explore any predictive benefits of a bottom-up deterministic pass of the inference network, especially in the case of natural images,  combined with optimization alternatives in order to prevent posterior collapse. Moreover, dividing the training data across the uncoupled encoders in the hierarchy could further speed-up inference and training. Finally, integration of pixel-regressive decoders and importance-weighted variations of the proposed scheme constitute directions for future research.}
\section*{Broader Impact}
Variational Autoencoders and other generative models have become widely influential in a variety of domains. VAEs 
have shown promise in modeling brain functions \cite{pinaya2019using,han2019variational}. They are also increasingly becoming a computational tool in network representation \cite{choong2018learning,yang2020relation}.
Recently, VAEs have been used for data integration in bioinformatics \cite{simidjievski2019variational,riesselman2018deep,berchuck2019estimating} and healthcare \cite{zhang2018generative,deng2019collaborative,yue2020deep}.

Moreover, there has been a surge in leveraging Variational Autoencoders for data compression \cite{golinski2020feedback}. The experimental results presented in Section \ref{sec: self_refl_mlp_experiments} are a strong indicator that the proposed architecture can realize smaller reconstruction loss with fewer latent dimensions (100 vs 128 latent variables). Compared to other computationally efficient alternatives, such as the LVAE, this would be particularly impactful in scenarios where inference time is critical such as online compression.

All real-world---and often very large---systems make the design of efficient inference algorithms imperative but difficult. Currently, the largest obstacle is scaling these systems to high-dimensional latent spaces. These properties are partly dictated by the assumptions on the variational distribution used. Therefore, rectifying these assumptions towards greater representational capacity, without compromising computational efficiency, becomes of utmost importance. We believe our work serves as a step in this desirable direction.



\bibliographystyle{unsrt}
\bibliography{neurips_2020}
\newpage
\section{Experimental Details of the MLP-SeRe VAE for binarized MNIST}
This section provides detailed description of the training parameters and architectural hyperparameters for the experiments in Section \ref{sec: self_refl_mlp_experiments} in the main paper.

Specifically, in Table \ref{tab:mlp_sere_training}, we provide the parameters of the training procedure. A constant learning rate and a small amount of weight regularization was used. We did not observe overfitting. Moreover, batch normalization layers were added at the input of each layer in the hierarchy.

\begin{table}[ht]
\centering
    \caption{Training Hyperparameters of the MLP-SeRe VAE's for the binarized MNIST}\label{tab:mlp_sere_training}
    \begin{tabular}{@{} p{4cm} *{2}{>{\ttfamily}l} @{}}
    \toprule[.1em]
    \normalfont Parameter & \normalfont Value \\
    \midrule[.1em]
    batch size & 256 \\
    warm up epochs &  1024 \\
    warm up schedule & geometric, $N=10$ KL levels \\
    epochs & 4000 \\
    learning rate & 1e-3\\
    batch normalization & Yes \\
    kernel/bias regularization & $\ell_2$, $\lambda=1e-5$ \\
    kernel/bias initializer & glorot normal\\
  \bottomrule[.1em]
  \end{tabular}
\end{table}

The model consists of 10 layers of 10 latent variables each. This experiment uses exclusively MultiLayer Perceptrons (MLP) as building blocks of the prior, posterior and transformational layers and of the final data distribution in the decoder.
The hyperparameters for each component in the hierarchy are given in Table \ref{tab:mlp_sere_vae}.
The evidence encoders of each layer are decoupled: they receive the raw binary image as input and not the output of the encoders at the upper (top-down inference) or lower (bidirectional inference \cite{kingma2016improved}) layer in the hierarchy. We also use \textit{latent encoders}, for all but the first inference layer in the hierarchy. These components process the latent codes provided by the bijective layer at the upper layer, before it is passed to the variational layer. A concatenation of the processed latent codes (output of the latent encoder) and the dataset feature map (output of the evidence encoder), are passed to the networks for learning the mean and the scale of the diagonal Gaussian distribution. In contrast to the parametrization adopted in Equations (13) and (14) of \cite{kingma2016improved} for the mean and variance, which restricts the scale in $(0,1)$ to ensure training stability, we use the following alternative that was found to be both more flexible and stable:
\begin{align}
& \sigma^2=softplus(elu(\Sigma_{out})),
\label{eq:gaussian_scale}
\end{align}
where $\Sigma_{out}$ is the network responsible for learning the scale of the distribution.
According to \eqref{eq:gaussian_scale}, large positive entries are left unaffected, while negative outputs of $\Sigma_{out}$ are first suppressed by the elu activation, and then mapped to a small positive value through the softplus transformation. A small offset is added to the small positive entries by the softplus to discriminate them by the negative outputs. A similar parametrization is used for the scale of the diagonal plus unit-rank affine transformations (to ensure positivity of the diagonal part and hence invertibility of the resulting bijective function).\\\\\\\


\begin{table}[ht]
\centering
    \caption{Architectural Hyperparameters of the MLP-SeRe VAE's layers for the binarized MNIST}\label{tab:mlp_sere_vae}
    \begin{tabular}{@{} p{8cm} *{2}{>{\ttfamily}l} @{}}
    \toprule[.1em]
    Component & \normalfont Parameter & \normalfont Value \\
    \midrule[.1em]
      \multirow{8}{*}{Evidence Encoder} & \# hidden layers  & 2 \\
      \addlinespace 
      & hidden dimension  & 256 \\
      \addlinespace 
      & feature size & 20 \\   
      \addlinespace
      & activation & Relu \\ 
      \addlinespace
      & output activation & None \\ 
      \midrule[.1em] 
            \multirow{8}{*}{Latent Encoder} & \# hidden layers  & 2 \\
      \addlinespace 
      & hidden dimension  & 256 \\
      \addlinespace 
      & feature size & 20 \\   
      \addlinespace
      & activation & Relu \\ 
      \addlinespace
      & output activation & None \\ 
      \midrule[.1em] 
       \multirow{3}{*}{\shortstack[l]{Variational Layer (diagonal Gaussian)\\2 identical networks (for loc and scale\_diag) \\
    MultivariateNormalDiag in \cite{tfptensorflow}}}
      & \# hidden layers  & 2 \\
      \addlinespace 
       & hidden dimension  & 256 \\
      \addlinespace 
       & activation & Relu \\ 
      \addlinespace
       & output activation & None \\ 
    \midrule[.1em] 
      \multirow{3}{*}{\shortstack[l]{Bijective Layer (diagonal plus unit-rank affine) \\3 identical networks \\ (for shift, scale\_diag, and scale\_perturb\_factor) \\
    Affine in \cite{tfptensorflow}}}
      & \# hidden layers  & 2 \\
      \addlinespace 
       & hidden dimension  & 20 \\
      \addlinespace 
       & activation & tanh \\ 
      \addlinespace
       & output activation & tanh \\ 
    \midrule[.1em] 
    \multirow{3}{*}{\shortstack[l]{Prior Layer (diagonal Gaussian) \\ 2 identical networks (for loc and scale\_diag) \\
    MultivariateNormalDiag in \cite{tfptensorflow}}}
      & \# hidden layers  & 2 \\
      \addlinespace 
       & hidden dimension  & 256 \\
      \addlinespace 
       & activation & Relu \\ 
      \addlinespace
       & output activation & None \\
    \midrule[.1em] 
    \multirow{2}{*}{\shortstack[l]{{Decoder} \\
    logit-based parametrization of Bernoulli in \cite{tfptensorflow}}}
      & \# hidden layers  & 2 \\
      \addlinespace 
       & hidden dimension  & 512 \\
      \addlinespace 
       & activation & Relu \\ 
      \addlinespace
       & output activation & None \\ 
\bottomrule[.1em]
\end{tabular}
\end{table}
\newpage
\section{Experimental Details of the ResNet-SeRe VAE for binarized MNIST}
This section provides detailed description of the training parameters, Table \ref{tab:resnet_sere_train} and architectural hyperparameters, Table \ref{tab:resnet_hyperparams}, for the experiments in Section \ref{sec: self_refl_resnet_experiments} in the main paper.

In contrast to the MLP-based architecture, where the two conditioning streams of the variational layer were concatenated as a single input for the networks of its distributional parameters, we use a more sophisticated structure, shown in Figure \ref{fig:residual_var_layer} in order to handle the unbalanced sizes of the latent and evidence feature maps. Specifically, we learn \textit{two separate} posterior networks that are connected in a \textit{residual manner}, as explained below. The first network is conditioned on the latent codes of the previous transformational layer, yielding a first \textit{estimation} $\phi^l_1(\boldsymbol{z}^{l-1})$ of the posterior parameters. The second layer is conditioned on the feature map provided by the ResNet encoder, and a hidden feature map that downscales the first estimation, and it actually learns a residual function $r^l(\boldsymbol{z}^{l-1},\boldsymbol{\mathcal{D}})$ that \textit{rectifies}
the first estimation, such that the final parameters are given by:
\begin{align*}
    \phi^l(\boldsymbol{z}^{l-1},\boldsymbol{\mathcal{D}})=\phi^l_1(\boldsymbol{z}^{l-1})+r^l(\boldsymbol{z}^{l-1},\boldsymbol{\mathcal{D}}).
\end{align*}
Intuitively, the inference consists of two steps. It uses latent information, that is a product of ``communication'' between the previous prior and posterior beliefs (output of the shared bijective layer), to obtain a first estimate of the relevant parameters. This estimate undergoes further refinement subject to the evidence presented to the variational layer. In our experiment, the networks for providing the hidden feature map of $\phi^l_1(\boldsymbol{z}^{l-1})$ have the same hyperparameters with the networks of $\phi^l_1(\boldsymbol{z}^{l-1})$ and output size $\floor*{\frac{D^l}{3}}$, where $D^l$ the dimension of the latent space at level $l$ (here $D^l=10$).

For the ResNet encoders and decoder, we use ResNet blocks, with batch normalization layers between them, that follow the design rule suggested in \cite{He_2016_CVPR}: if the feature map size is halved, the number
of filters is doubled, and reversely if the feature map size is doubled the number of filters is halved, so as to preserve the time complexity per layer. This corresponds to feature maps $[16,32]$ in the encoder (uncoupled encoders of two ResNet blocks, with donwnsampling at both) and $[32,16]$ in the decoder (two ResNet blocks with transpose convolution). In the encoder, there is one more fully-connected layer after the convolutional layers, with no hidden units, yielding the final feature map of size $64$.

\begin{figure*}
\centering
\includegraphics[height=2.5in]{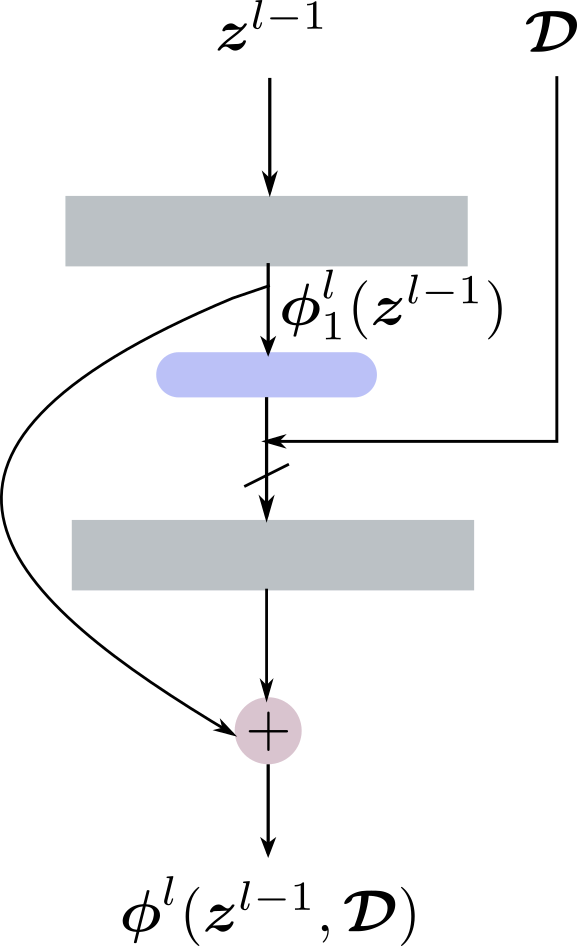}
\caption{{\small
\textit{Building Block of the ResNet Variational Layer with  a residual connection for combining $\boldsymbol{z}^{l-1}$ and $\boldsymbol{\mathcal{D}}$.}} A residual connection is used for the computation of the variational parameters $\boldsymbol{\phi}^l$ of the $l$-th latent layer between the $\boldsymbol{\phi}^l$ conditioned by the previous latent factors $\boldsymbol{z}^{l-1}$ and the $\boldsymbol{\phi}^l$ conditioned by the evidence $\boldsymbol{\mathcal{D}}$.}
\label{fig:residual_var_layer}
\end{figure*}
\newpage
\begin{table}[ht]
\centering
    \caption{Architectural Hyperparameters of the ResNet-SeRe VAE's layers for the binarized MNIST}\label{tab:resnet_hyperparams}
    \begin{tabular}{@{} p{8cm} *{2}{>{\ttfamily}l} @{}}
    \toprule[.1em]
    Component & \normalfont Parameter & \normalfont Value \\
    \midrule[.1em]
      \multirow{8}{*}{Evidence Encoder} & initial $\#$ filters  & 16  \\
      \addlinespace 
      & $\#$ ResNet blocks  & 2 \\
       \addlinespace 
      & ResNet blocks scale  & [$\downarrow 2$,$\downarrow 2$] \\
      \addlinespace 
      & feature size & 64 \\   
      \addlinespace
      & kernel size & 3 \\   
      \addlinespace
      & activation & Relu \\ 
      \addlinespace
      & output activation & None \\ 
      \midrule[.1em] 
            \multirow{8}{*}{Latent Encoder} & \# hidden layers  & 2 \\
      \addlinespace 
      & hidden dimension  & 256 \\
      \addlinespace 
      & feature size & 20 \\   
      \addlinespace
      & activation & Relu \\ 
      \addlinespace
      & output activation & None \\ 
      \midrule[.1em] 
       \multirow{3}{*}{\shortstack[l]{Variational Layer (diagonal Gaussian) \\2 identical networks (for loc and scale\_diag) \\
    MultivariateNormalDiag in \cite{tfptensorflow}}}
      & \# hidden layers  & 2 \\
      \addlinespace 
       & hidden dimension  & 256 \\
      \addlinespace 
       & activation & Relu \\ 
      \addlinespace
       & output activation & None \\ 
      \addlinespace
         & hidden feature size & 3 \\ 
    \midrule[.1em] 
      \multirow{3}{*}{\shortstack[l]{Bijective Layer (diagonal plus unit-rank affine)\\3 identical networks \\ (for shift, scale\_diag, and scale\_perturb\_factor) \\
    Affine in \cite{tfptensorflow}}}
      & \# hidden layers  & 2 \\
      \addlinespace 
       & hidden dimension  & 20 \\
      \addlinespace 
       & activation & tanh \\ 
      \addlinespace
       & output activation & tanh \\ 
    \midrule[.1em] 
    \multirow{3}{*}{\shortstack[l]{Prior Layer (diagonal Gaussian) \\ 2 identical networks (for loc and scale\_diag) \\
    MultivariateNormalDiag in \cite{tfptensorflow}}}
      & \# hidden layers  & 2 \\
      \addlinespace 
       & hidden dimension  & 256 \\
      \addlinespace 
       & activation & Relu \\ 
      \addlinespace
       & output activation & None \\
    \midrule[.1em] 
    \multirow{2}{*}{\shortstack[l]{{Decoder} \\
    logit-based parametrization of Bernoulli in \cite{tfptensorflow}}} &
     initial \# filters  & 32  \\
      \addlinespace 
      & \# ResNet blocks'  & 2 \\
       \addlinespace 
      & ResNet blocks' scale  & [$\uparrow 2$, $\uparrow 2$] \\
      \addlinespace
      & kernel size & 3 \\   
      \addlinespace
      & activation & Relu \\ 
      \addlinespace
      & output activation & None \\ 
\bottomrule[.1em]
\end{tabular}
\end{table}
\begin{table}[ht]
\centering
    \caption{Training Hyperparameters of the ResNet-SeRe VAE's for the binarized MNIST}
    \begin{tabular}{@{} p{4cm} *{2}{>{\ttfamily}l} @{}}
    \toprule[.1em]
    \normalfont Parameter & \normalfont Value \\
    \midrule[.1em]
    batch size & 128 \\
    warm up epochs &  256 \\
    warm up schedule & linear \\
    epochs & 1000 \\
    learning rate & 1e-3\\
    batch normalization & Yes \\
    kernel/bias regularization & No \\
    kernel/bias initializer & glorot normal\\
    \bottomrule[.1em]
  \end{tabular}\label{tab:resnet_sere_train}
\end{table}
\newpage
\section{Experimental Details of the variational SeRe MAF for CIFAR-10}
This section provides detailed description of the training parameters, Table \ref{tab:train_maf}, and architectural hyperparameters Table \ref{tab:maf_hyperparams}, for the experiments in Section \ref{sec: self_refl_maf_experiments} in the main paper. The model consists of 5 layers of 40 latent variables each.

\subsection{Training set-up}
Table \ref{tab:train_maf} summarizes the hyperparameters of the training process. Below, we describe techniques adopted that are different than those used for the models in the previous experimental studies.
\begin{itemize}
\item Due to the complexity of the decoder, which stems from the autoregressive property and the large hidden dimension required in the autoregressive layer, we follow a 3-phase training scheme. First, we minimize only the reconstruction loss (the negative conditional likelihood) of the negative ELBO objective loss, Equation \eqref{eq:regularized_elbo}, for 350 epochs. Subsequently, we apply geometric warm-up for $N=8$ KL levels (and other 258 epochs). We, finally optimize the rest of the ELBO, for the remaining epochs. For the first phase, we use adaptive learning rate with a cosine scheduler. For the last two phases, a constant learning rate is used.
\item The pixel space is converted to logit space by the transformation: $\boldsymbol{x} \rightarrow logit(\alpha+(1-\alpha)x)$, for $\alpha=0.05$, after adding uniform noise and  rescaling the pixel values in $[0,1]$, as in \cite{papamakarios2017masked}.
\item We find applying dropout \cite{srivastava2014dropout}, with probability $0.5$, in the residual blocks of this architecture beneficial for further regularization.
\item We employ a \textit{free bits} strategy, as described in the Appendix C.8, Equation (12), in \cite{kingma2016improved}. In our case, the latent variables are grouped by layer (so that $K$ in Equation (12) in \cite{kingma2016improved} equals $L=4$) with $\lambda_{free_\_bits}=1$.
\end{itemize}

\begin{table}[ht]
\centering
    \caption{Training Hyperparameters of the variational SeRe MAF for CIFAR-10}
    \begin{tabular}{@{} p{6cm} *{2}{>{\ttfamily}l} @{}}
    \toprule[.1em]
    \normalfont Parameter & \normalfont Value \\
    \midrule[.1em]
    batch size & 512 \\
    warm up epochs &  350+256 \\
    warm up schedule & hard (350 epochs), geometric (258 epochs for $N=8$) \\
    epochs & 1500 \\
    free bits & 1 \\
    learning rate & cosine ([1,350] epochs), $1e-3$ ([351,1500])\\
    batch normalization & Yes \\
    kernel/bias regularization & $\ell_2$, $\lambda=1e-3$ \\
    kernel initiralizer & glorot normal \\
   $\alpha$ for logit-space as in \cite{papamakarios2017masked} & 0.05 \\
   \bottomrule[.1em]
  \end{tabular}\label{tab:train_maf}
\end{table}
\newpage
\subsection{Conditional Masked AutoEncoder}
In this section, we describe the construction of the \textit{conditional Masked Autoencoders} (MADE layers) used as building blocks for the variational normalizing flow presented in Section \ref{sec: self_refl_maf_experiments}. We use notation identical to those used in \cite{germain2015made}.

Let $C$ be the dimension of the conditioning inputs. $C$ acts as a \textit{mask offset} in the construction of the masked autoregressive encoder, as we explain below. We assign unique numbers $1,2,\dots,C+D$ to the inputs. In case of a random input ordering, the first $C$ conditioning inputs are excluded so that $m^0(d)=d$, for $d=1,2,\dots,C$ and $m^0(d)\in\{C+1,C+2,\dots,C+D\}$ uniquely and randomly assigned to the inputs $d=C+1,C+2,\dots,C+D$. The degrees $m^l(d)$ of the $d-$th hidden unit of layer $l$ should now be larger than $C$, so that the conditioning inputs are not masked out: the conditioning inputs are connected to all the hidden units. Therefore, $m^l(d)$ are random numbers such that $m^l(d)\in \{C+1,C+2,C+D\}$.
Equation (12) in \cite{germain2015made} is still valid for the construction of the masks for connections from the input to the first layer hidden units, and from hidden units to next layer hidden units. For the last layer masks (from the hidden units to the output), Equation (13) in \cite{germain2015made} is used, and subsequently the first $C$, that refer to the conditioning inputs, masks are discarded. 

Finally, as suggested in \cite{papamakarios2017masked} batch normalization layers between the MAF steps are incorporated. Section B in the Appendix of \cite{papamakarios2017masked}, provides a description of the batch normalization as a bijector and $tfp.bijectors.BatchNormalization$ \cite{tfptensorflow} offers a suggested implementation. In our implementantion, at both training and validation/test time, we maintain averages over minibatches as in \cite{pmlr-v37-ioffe15}.

\begin{longtable}{@{} p{8cm} *{2}{>{\ttfamily}l} @{}}
    \caption{Architectural Hyperparameters of the variational SeRe MAF for CIFAR-10}\label{tab:maf_hyperparams}\\
    \toprule[.1em]
    Component & \normalfont Parameter & \normalfont Value \\
    \midrule[.1em]
      \multirow{8}{*}{Evidence Encoder} & initial \# filters  & 16  \\
      \addlinespace 
      & \# ResNet blocks  & 3 \\
       \addlinespace 
      & ResNet blocks' scale  & [$\downarrow 2,\downarrow 2,\downarrow 2$] \\
      \addlinespace 
      & feature size & 128 \\   
      \addlinespace
      & kernel size & 3 \\ 
      \addlinespace
      & dropout probability & 0.5 \\ 
      \addlinespace
      & activation & Relu \\ 
      \addlinespace
      & output activation & None \\ 
      \midrule[.1em] 
            \multirow{8}{*}{Latent Encoder} & \# hidden layers  & 2 \\
      \addlinespace 
      & hidden dimension  & 100 \\
      \addlinespace 
      & feature size & 80 \\   
      \addlinespace
      & activation & Relu \\ 
      \addlinespace
      & output activation & None \\ 
      \midrule[.1em] 
       \multirow{3}{*}{\shortstack[l]{Variational Layer (diagonal Gaussian) \\2 identical networks (for loc and scale\_diag) \\
    MultivariateNormalDiag in \cite{tfptensorflow}}}
      & \# hidden layers  & 1 \\
      \addlinespace 
       & hidden dimension  & 512 \\
      \addlinespace 
       & activation & Relu \\ 
      \addlinespace
       & output activation & None \\ 
    \midrule[.1em] 
    \multirow{2}{*}{\shortstack[l]{Variational Layer - \\hidden feature maps for the posterior distribution \\
    2 identical networks \\
    (for the residual connections of loc, scale\_diag)
 }} 
    & $\#$ hidden layers & 1 \\
     \addlinespace 
    & hidden dimension & 256 \\
     \addlinespace 
    & feature size & 40  \\
      \addlinespace 
    & activation & Relu \\ 
      \addlinespace
    & output activation & None \\ 
    \midrule[.1em] 
      \multirow{3}{*}{\shortstack[l]{Bijective Layer (diagonal plus unit-rank affine) \\3 identical networks \\ (for bin\_widths, bin\_heights, and knot\_slopes) \\
    RationalQuadraticSpline in \cite{tfptensorflow}}}
      & \# hidden layers  & 2 \\
      \addlinespace 
       & hidden dimension  & 60 \\
      \addlinespace 
       & activation & tanh \\ 
      \addlinespace
       & output activation & tanh \\ 
      \addlinespace
      & \# bins & 32 \\
      \addlinespace
      & \# splines & 5 \\ 
      \addlinespace
      & mask size ($d$ in \cite{durkan2019neural}) & 5 \\ 
      \addlinespace
      & range\_min ($B$ in \cite{durkan2019neural}) & -20 \\ 
    \midrule[.1em] 
    \multirow{3}{*}{\shortstack[l]{Prior Layer (diagonal Gaussian) \\ 2 identical networks (for loc and scale\_diag) \\
    MultivariateNormalDiag in \cite{tfptensorflow}}}
      & \# hidden layers  & 1 \\
      \addlinespace 
       & hidden dimension  & 256 \\
      \addlinespace 
       & activation & Relu \\ 
      \addlinespace
       & output activation & None \\
    \midrule[.1em] 
    \multirow{2}{*}{\shortstack[l]{{Decoder - base distribution (unit rank Gaussian)} \\
    3 identical networks  \\
    (for loc, scale\_diag,scale\_perturb\_factor)\\ MultivariateNormalDiagPlusLowRank in \cite{tfptensorflow}}} &
     initial \# filters  & 64  \\
      \addlinespace 
      & \# ResNet blocks'  & 3 \\
       \addlinespace 
      & ResNet blocks' scale  & [$\uparrow 2, \uparrow 2,\uparrow 2$] \\
      \addlinespace
      & kernel size & 3 \\   
      \addlinespace
      & dropout probability & 0.5 \\
      \addlinespace
      & activation & Relu \\ 
      \addlinespace
      & output activation & None \\ 
    \midrule[.1em] 
    \multirow{2}{*}{\shortstack[l]{{Decoder - hidden feature maps for the base distribution } \\
    3 identical networks  \\
    (for the residual connections of:\\ 
    loc, scale\_diag,scale\_perturb\_factor)}}\\ 
    & \# hidden layers & 2 \\
     \addlinespace 
    & hidden dimension & 512 \\
     \addlinespace 
    & feature size  & 100  \\
      \addlinespace 
    & activation & Relu \\ 
      \addlinespace
    & output activation & None \\ 
    \midrule[.1em] 
    \multirow{3}{*}{\shortstack[l]{Decoder - autoregressive bijector (MAF)}}
      & \# flows  & 2 \\
       \addlinespace 
      & \# MADEs/flow  & 5 \\
       \addlinespace 
      & batch normalization & Yes \\ 
       \addlinespace 
      & MADE: \# hidden layers  & 2 \\
      \addlinespace 
       & MADE: hidden dimension  & 1024 \\
      \addlinespace 
       & MADE: activation & Relu \\ 
      \addlinespace
       & MADE: output activation & None \\
      \addlinespace
       & MADE: input order & random \\
      \addlinespace
       & MADE: hidden degrees & equal \\ 
\bottomrule[.1em]
\end{longtable}

The trained models can be downloaded from: \\
\url{https://drive.google.com/file/d/1W_wBTCOM8FouxlJT2IbSYvuwPhpkfpcm/view?usp=sharing}




\end{document}